\documentclass[accepted]{uai2021} % after acceptance, for a revised
\usepackage[american]{babel}
\usepackage{natbib} % has a nice set of citation styles and commands
\usepackage{mathtools} % amsmath with fixes and additions
\usepackage{booktabs} % commands to create good-looking tables
\usepackage{tikz} % nice language for creating drawings and diagrams

\usepackage{microtype}
\usepackage{graphicx}
\usepackage{subcaption}
\usepackage{amssymb}
\usepackage[ruled,vlined]{algorithm2e}
\usepackage{xcolor}
\usepackage{multirow}
\usepackage{outlines}

\usepackage{hyperref}

\DeclareMathOperator{\Relu}{ReLU}
\DeclareMathOperator{\Proj}{Proj}
\DeclareMathOperator{\Inv}{Inv}
\DeclareMathOperator{\Tr}{Tr}
\DeclareMathOperator{\argmax}{argmax}
\DeclareMathOperator{\argmin}{argmin}
\newcommand{\Exp}[2]{\mathbb{E}_{#1} #2}
\newcommand{\R}{\mathbb{R}}
\newcommand{\relu}[1]{\Relu\left( #1 \right)}
\newcommand{\T}{\mathsf{T}}
\newcommand{\calL}{\mathcal{L}}

\newcommand{\calN}{\mathcal{N}}

\newcommand{\wmw}{\begin{bmatrix} W \\ -W \end{bmatrix}}
\newcommand{\pmw}[1]{\begin{bmatrix} #1 \\ -#1 \end{bmatrix}}
\newcommand{\pmwt}[1]{\begin{bmatrix} #1\ -#1 \end{bmatrix}}
\newcommand{\bm}[2]{\begin{bmatrix} #1\\ #2 \end{bmatrix}}

\newcommand{\lmse}{\calL_\text{MSE}}
\newcommand{\lml}{\calL_\text{ML}}
\newcommand{\inflow}{\textsc{Trumpet}}
\newcommand{\iflow}{iFlow}

\usepackage{amsthm}
\newtheorem{prop}{Claim}

\title{\textsc{Trumpets}: Injective Flows for Inference and Inverse Problems}

\author[1]{\href{mailto:Konik Kothari <kkothar3@illinois.edu>?Subject=Your UAI 2021 paper}{Konik~Kothari}{}} % Lead author
\author[2]{AmirEhsan Khorashadizadeh}
\author[3]{Maarten de Hoop}
\author[2]{Ivan Dokmani\'c}

\affil[1]{%
    Coordinated Science Laboratory\\
    University of Illinois at Urbana-Champaign
}
\affil[2]{%
    Department of Mathematics and Computer Science\\
    University of Basel
}
\affil[3]{ 
Computational and Applied Mathematics\\
Rice University}

\begin{document}
\maketitle
\begin{abstract}
We propose injective generative models called {\inflow s}~that generalize invertible normalizing flows. The proposed generators progressively increase dimension from a low-dimensional latent space. We demonstrate that {\inflow s} can be trained orders of magnitudes faster than standard flows while yielding samples of comparable or better quality. They retain many of the advantages of the standard flows such as training based on maximum likelihood and a fast, exact inverse of the generator. Since {\inflow s} are injective and have fast inverses, they can be effectively used for downstream Bayesian inference. To wit, we use {\inflow} priors for maximum a posteriori estimation in the context of image reconstruction from compressive measurements, outperforming competitive baselines in terms of reconstruction quality and speed. We then propose an efficient method for posterior characterization and uncertainty quantification with {\inflow s} by taking advantage of the low-dimensional latent space.
\end{abstract}

% \section{Outline}
% \begin{outline}
% \1 Introduction
%     \2 Unsupervised learning, generative modeling
%     \2 Advantages of flows
%     \2 Disadvantages of flows
%     \2 Injective flows -> Our contribution
%     \2 Computational efficiency
%     \2 Downstream applications

% \1 Injective flows 
%     \2 Define flows
%     \2 Making flows injective
%         \3 Standard coupling layers
%         \3 Injective coupling layers (\*)
%             \4 Figure to illustrate? (not A-priority)
%             \4 Fast inverses
%             \4 Stability proposition
%     \2 InFlows - A generative model based on injective coupling layers
%         \3 Complete architecture (short)
%         \3 MSE + ML training
%     \2 Inference with InFlows
%         \3 Image reconstruction + Bayes + MAP
%         \3 Exact likelihoods vs bounds
%             \4 Stochastic approximation
%             \4 To solve inference problems like MAP only need gradient
%             \4 Claim that some of this is deferred to the appendix
        
% \1 Experiments
%     \2 Setup
%     \2 Generative performance
%     \2 Inverse problems + MAP
% \end{outline}        

% \tableofcontents

\section{Introduction}

Modeling a high-dimensional distribution from samples is a fundamental task in unsupervised learning. An ideal model would efficiently generate new samples and  assign likelihoods to existing samples. Some deep generative models such as generative adversarial networks (GANs)~\citep{goodfellow2014generative} can produce samples of exceedingly high quality, but they do not give access to the underlying data distribution. Moreover, GANs are often hard to train, suffering from pathologies such as mode collapse~\citep{thanh2020catastrophic,arjovsky2017towards}. Since they are generally not invertible, or computing the inverse is slow, they are not well-suited for downstream inference tasks such as image reconstruction from compressive measurements or uncertainty quantification.

Normalizing flows alleviate many of the drawbacks of GANs: they approximate high-dimensional probability distributions as invertible transformations of a simple, tractable base distribution. They allow both efficient sampling and likelihood evaluation. They can be trained using maximum likelihood, 
% they do not seem to suffer from mode collapse, 
and at inference time they provide direct access to likelihoods. These desirable features are a consequence of clever architectural components known as coupling layers~\citep{dinh2014nice}. 

Normalizing flows, however, are extremely compute-intensive. As a case in point, training a Glow model \citep{kingma2018glow} for the 5-bit $256\times256$ CelebA dataset takes a week on 40 GPUs. This is in part because the dimension of the ``latent'' space in normalizing flows equals that of the generated images. Since signals of interest are often concentrated close to low-dimensional structures embedded in high-dimensional spaces, this is a waste of resources. Beyond reducing computational cost, a low-dimensional latent space acts as a natural regularizer when solving ill-posed inverse problems~\citep{bora2017compressed}.

In this paper we propose a new generative model termed \inflow---an injective flow based on convolutional layers that are \textit{injective} by construction. Similarly to traditional coupling layers our proposed layers have fast, simple inverses and tractable Jacobians; however, they map to a space of higher dimension. Since they are injective, they can be inverted on their range. Our design combines standard coupling layers with recent results on injective neural networks \citep{puthawala2020globally}. Further, our models can be trained via exact maximum likelihood by separating the training of the injective part from that of the bijective part \citep{brehmer2020flows}.

% The same strategy has been used very recently to design injective generative models which use traditional coupling layers.  

{\inflow s} can be trained orders of magnitude faster than previous injective models based on traditional normalizing flows ~\citep{brehmer2020flows} while producing samples of comparable (or better) quality. Moreover, thanks to their fast inverse, they can be used to design fast inference algorithms based on generative priors. We apply {\inflow s} to Bayesian inference problems in compressive sensing and limited-angle tomography. In particular, we devise an algorithm for efficient computation of a MAP estimator using a variant of projected gradient descent. The projection is computed via the fast inverse while thanks to injectivity we can access the likelihoods. We then adapt recent work on uncertainty quantification for inverse problems with normalizing flows \citep{sun2020deep} to work with generative priors and a low-dimensional latent space of {\inflow s}. We anticipate that neural-network-based uncertainty quantification can be naturally integrated in a rigorous analysis in the context of inverse problems \citep{mosegaard1995monte,monard2020consistent}.
% \md{The perspective of sampling a posterior distribution for ``solving'' inverse problems has a long history; we mention an original paper by Mosegaard and Tarantola [please add the reference].}  
% \id{The Monard reference is rather for ``principled'' UQ for inverse problems; perhaps to conclusion or to related work?}\kk{can you add this? I am not sure which one you refer to...}

Our \textbf{main contributions} can be summarized as follows:
\begin{itemize}
    \item We propose \textit{injective} coupling layers with fast inverses and tractable Jacobians.
    \item We use these layers to construct {\inflow s}---injective flow generative models. The proposed generative models train orders of magnitude faster than the usual flow models while producing samples of comparable or better quality and giving access to likelihoods.
    \item We apply the proposed models to Bayesian inference problems and uncertainty quantification, showing remarkable gains in efficiency as well as reconstruction quality over established methods. In particular, we show how the low-dimensional latent space of \inflow s leads to an efficient variational approximation of the posterior distribution. 
    % \id{Mention favorable sample complexity of MCMC in low-dim space?}
\end{itemize}

In the following section we describe the construction of {\inflow s}; an overview of related work is given in Section~\ref{sec:related-work}.

\section{{\inflow s}: Injective flows}

\begin{figure*}
\centering
\includegraphics[width=0.8\textwidth]{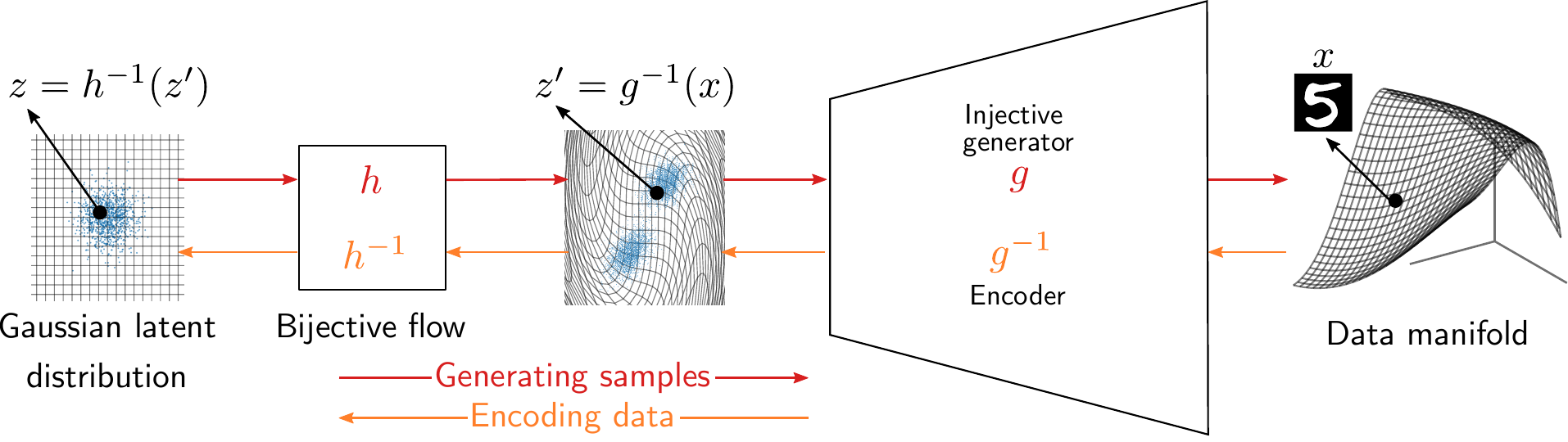}
\caption{\inflow---A reversible injective flow-based generator}
\label{fig:network_arch}
% \md{[perhaps we could illustrate here also the generation of a multimodal distribution]}

\end{figure*}

Flow-based generative models~\citep{dinh2014nice,dinh2016density} approximate the target distribution via a series of bijective transformations of a simple latent distribution. Unlike GANs~\citep{goodfellow2014generative} or VAEs~\citep{kingma2013auto} they allow for efficient \emph{exact} likelihood evaluation. Crucial to the design of flow-based models are tractable inverses and Jacobians of all the constituent bijective transformations~\citep{kingma2018glow,grathwohl2018ffjord}, based on special coupling layers such as NICE~\citep{dinh2014nice} or Real-NVP \citep{dinh2016density}. A generative model $f_\theta : \R^D \to \R^D$ parameterized by the weights $\theta$ maps latent variables $Z$ to data $X$. Note that we use uppercase letters for random vectors and corresponding lowercase letters for their realizations. Log-likelihoods of the generated samples $x = f_\theta(z)$ can be evaluated as
\begin{equation} \label{eq:cov}
\log p_X(x) = \log p_Z(f_\theta^{-1}(x)) - \log |\det J_{f_\theta}(f_\theta^{-1}(x))|.
\end{equation}
Given an iid training dataset $\{\xi^{(i)}\}_{i=1}^n$ from some ground truth distribution\footnote{We use $\xi$ to denote samples from the ground truth distribution $p_\Xi$ to distinguish them from the samples $x$ from $p_X$, the distribution induced by our network $f_\theta$.} $p_{\Xi}$, training a normalizing flow entails maximizing the log-likelihood of the training data given as $\sum_{i=1}^{N}\log p_X(\xi^{(i)})$ over the weights $\theta$ in order to learn a generative model $f_\theta$. Equivalently, it entails minimize the KL divergence between $p_X$ and $p_\Xi$. While invertibility ensures a non-singular $J_{f_\theta}$ at all points, defining likelihoods only requires injectivity of $f_{\theta}$.

\subsection{Making flows injective}
\label{sec: injective_flow}

Machine learning for high-dimensional signals such as images relies on the fact that these signals concentrate around low-dimensional structures. We adopt the common assumption that $p_\Xi$ is concentrated close to a $d$-dimensional manifold in $\R^D$, with $d \ll D$. We then aim to learn a generative model $f_\theta$, now mapping from $\R^d$ to $\R^D$, such that the observed data lies in the range of $f_\theta$. When $f_\theta$ is an injective map its Jacobian $J_{f_\theta} \in \R^{D\times d}$ has full column rank for all input points. Thus one can still have access to likelihoods of samples generated by $f_\theta$ by modifying \eqref{eq:cov} as~\citep{boothby1986introduction}
\begin{multline}
\label{eq:cov_inj}
\log p_X(x) = \log p_Z(f_\theta^{\dagger}(x))
\\
- \dfrac{1}{2}\log |\det [J_{f_\theta}(f_\theta^{\dagger}(x))^\T J_{f_\theta}(f_\theta^{\dagger}(x))]|
\end{multline}
which is valid for $x \in \mathrm{Range}(f_\theta)$. We use $f_\theta^\dagger$ to denote an inverse of $f_\theta$ on its range, that is $f_\theta^\dagger(f_\theta(z)) = z$. As described later, due to the way we construct $f_\theta^\dagger$, Equation \eqref{eq:cov_inj} corresponds to the likelihood of a projection of $x$ on the range of $f_\theta$ for $x \notin \mathrm{Range}(f_\theta)$.
% Therefore, an injective flow model would require training in two steps. \id{break here and introduce our layers} \kk{is this what you had in mind?} \id{I would not at all mention these projections and samples out of range here, but rather say that you now show how to design an injective flow layer; ``We propose novel injective revnet...'' and then describe training...}

Building on the general change of variable formula \eqref{eq:cov_inj}, we propose \inflow---a network architecture that is injective by construction. The network architecture (Figure \ref{fig:network_arch}) consists of a ``flat'' invertible part which maps $\R^d$ to $\R^d$ and an expanding injective part which maps $\R^d$ to $\R^D$, resembling its namesake in shape. Crucially, expansion is enabled via injective revnet steps~\citep{jacobsen2018revnet} generalizing the recently proposed Glow~\citep{kingma2018glow} layers.

% First, we train an injective map,
% % \md{[ambiguous notation, should we introduce a composition (or pullback) here: $f = h^* g$ ]} 
% $g$ from a low-dimensional latent space such that the resulting manifold goes through samples from training data. Second, we train a regular bijective flow model, $h$ in the latent space that morphs a simple distribution to the distribution of $g$-inverses of training data $\{g^\dagger(x)\}$. A sample is then generated as $f(z) = g(h(z))$. Our architecture is depicted in Figure \ref{fig:network_arch}. A similar break-up of training phase was recently also discovered in ~\citep{brehmer2020flows}. Crucially, as a consequence of this two-fold training process we do not require $J_g$ during the training process. However, we still need fast inverses of the $g$ during training. We propose novel injective revnet steps~\citep{jacobsen2018revnet} generalizing recently proposed Glow~\citep{kingma2018glow} architectures that allow us to design provably injective networks.

We begin by reviewing the revnet step. A forward (F) revnet step has 3 operations, each having a simple inverse (I):
\begin{enumerate}[leftmargin=*]
    \item activation normalization,
    \[
    \text{\textsc{F:}} ~ y = \dfrac{x-\mu}{\sigma} \quad \quad \quad \quad \quad
    \text{\textsc{I:}} ~ x = \sigma y + \mu
    \]
    \item $1\times1$ convolution with a kernel $w$,
    \[
    \text{\textsc{F:}} ~ y = \ell_w(x) = w\ast x \quad \quad  
    \text{\textsc{I:}} ~ x = w^{-1}\ast y
    \]
    \item affine coupling layer
    \begin{equation*}
    \begin{array}{llcrlcl}
        \text{\textsc{F:}} & y_1 &\hspace{-2mm}=\hspace{-2mm}& x_1  & \quad y_2 &\hspace{-2mm}=\hspace{-2mm}& s(x_1) \circ x_2 + b(x_2)\\
        \text{\textsc{I:}} &x_1 &\hspace{-2mm}=\hspace{-2mm}& y_1  & \quad x_2 &\hspace{-2mm}=\hspace{-2mm}& s(y_1)^{-1} \circ (y_2 - b(y_1)),
        % \bm{y_1}{y_2} = \bm{x_1}{s(x_1) \odot x_2 + b(x_2)}
        % \leftrightarrow
        % \bm{x_1}{x_2} = \bm{y_1}{s(x_1)^{-1} \odot (y_2 - b(y_1)}
    \end{array}
    \end{equation*}
\end{enumerate}
where $y = \bm{y_1}{y_2}$ and $x = \bm{x_1}{x_2}$. 
Here $s$ and $b$ are the scale and bias functions respectively that are implemented by neural networks. The coupling layers have triangular Jacobians making their log determinants easy to compute.

We now generalize the second step to allow for an increase in dimension while retaining computational tractability.

\paragraph{Injective $1\times 1$ convolutions.}
We consider generalizations of the $1 \times 1$ convolution layers ($\ell_w$) that (1) are injective, (2) have fast (pseudo)inverse and (3) a fast Jacobian independent of $x$. These requirements yield two layer variants---linear and $\Relu$ $1\times 1$ convolutions:
% \id{the table floats uncontrollably---why not leave an equation array or something?} 
\begin{table}[h!] \label{tab:eqns}
\centering
\label{tab:inj_layers}
{%
\begin{tabular}{@{}lll@{}}
       & \textsc{Linear} & $\Relu$ \vspace{1mm} \\
\textsc{Forward} &   $y = w \ast x$      &  $y = \relu{ \pmw{w} \ast x}$       \vspace{2mm}\\
\textsc{Inverse}   &     $x := w^\dagger \ast y$    &    $x := w^\dagger \ast \left(\pmwt{I} y\right)$.     
\end{tabular}%
}
\end{table}

Here $w^\dagger$ is the left pseudoinverse of $w$. Since $w$ is a $1 \times 1$ convolution, we write it as a matrix of size $c_\text{out} \times c$, where $c$, $c_\text{out}$ are the number of input and output channels respectively; taking the pseudoinverse of this matrix yields $w^\dag$.

In Appendix \ref{app:derivations}, we show that for both types of layers,
\[ 
    \log\det J_{\ell_w}^\T J_{\ell_w} = \sum_{i=1}^{c} s_i(w)^2,
\]
where the $s_i(w)$ are the singular values of $w$. We choose the size of $w$ such that the number of output channels is $k c$ (resp. $\lfloor \frac{k}{2} \rfloor c$) for the linear (resp. ReLU) layer. While $k \ge 1$ is enough for the linear variant to be injective, $k \ge 2$ is necessary and sufficient for the $\Relu$ variant \citep{puthawala2020globally}.

\paragraph{Injective revnet step.}  By generalizing the 1$\times$1  convolutions to increase dimensions, we can still utilize the revnet step as in Glow by replacing the invertible $1 \times 1$ convolutions by their injective counterparts.
% \md{[I suggest to put the input/output through mapping properties in the text and use a similar format as in the description of the $1 \times 1$ convolutions above]}
% \id{Can you make this more readable without resizing? perhaps 3 columns: first one forward and inverse; second one actnorm injconv coupling, then again actnorm injconv coupling, third one the expressions?}

% \begin{align*}
%     &\textbf{Input}  &&x \in \R^{N\times N \times c}\\
%     &\textbf{Actnorm}  &&y = \dfrac{x-\mu}{\sigma}  \\
%     &\textbf{Injective conv} &&y = \ell_w(x)  \\
%     &\textbf{Affine coupling} &&\bm{y_1}{y_2} = \bm{x_1}{s(x_1) \odot x_2 + b(x_2)}\\
%     &\textbf{Output} &&y \in \R^{N\times N \times kc}
% \end{align*}

Therefore, if the input tensor is of size $N\times N\times C$, the output after an injective revnet step is of size $N\times N\times kC$, where the expansion by a factor $k$ occurs in the injective convolution $(\ell_w)$ step. 
% \id{What do you mean by a different context? I meant they used it for UQ and not image generation}

\subsection{Architecture of {\inflow s}} \label{sec:arch}

Injective coupling layers introduced in the previous section allow us to build an architecture that trains at a fraction of the time and memory cost of regular flows. As shown in Figure \ref{fig:network_arch}, a \inflow~model $f_\theta(z)=g_{\gamma}(h_{\eta}(z))$ with weights $\theta = (\gamma, \eta)$ has two components: an injective map $g_{\gamma}(z') =  g_1 \circ g_2 \ldots \circ g_K(z')$ which maps from $\R^d$ to $\R^D$, and a bijective part $h_{\eta}$ implemented as a flow $z' = h_\eta(z) = h_1 \circ h_2 \ldots \circ h_L(z)$ in the low-dimensional latent space.  Unlike normalizing flows such an architecture allows us to progressively increase dimension and markedly reduce the number of parameters.

The role of the injective part $g_\gamma$ is to match the shape of the manifold that supports the ground truth distribution $p_\Xi$, while the role of the low-dimensional flow is to match the density on the manifold. As we elaborate in Section \ref{sec:train}, and as was also recently noted by \cite{brehmer2020flows}, this separation enables training even when likelihood is not defined for samples outside the range of $f_\theta$.

To build the injective map $g_\gamma$ we compose the proposed injective revnet layers, progressively increasing dimension from that of the latent space to that of the image space. To improve expressivity, at each resolution, we interleave a small number of bijective revnet layers. Each injective layer increases feature dimension by a factor of $2$ in a single step in the forward direction (and decreases it by a factor of $2$ in the reverse direction). Following \cite{dinh2016density} we employ upsqueezing to increase resolution. Our network architecture results in significantly fewer parameters and faster training than the recently proposed variant of injective flows~\citep{brehmer2020flows}.
% Crucially, it allows for a low-dimensional latent space which acts as a natural regularizer in inverse problems (see Section \ref{sec:exp}).

Finally, performance of revnets in generative modeling of images can be improved~\citep{dinh2016density} by introducing multiscale implementations of the scale ($s$) and bias ($b$) functions. For these implementations, we propose to use U-Nets~\citep{ronneberger2015u} in affine coupling layers as opposed to regular convolutional stacks used in previous normalizing flows~\citep{dinh2016density,kingma2018glow}. We find that integrating U-Nets greatly improves the performance of our network.
% A related approach with fully connected layers was previously used by~\citet{sun2020deep}.

\subsection{Training of {\inflow s}} \label{sec:train}

An advantage of injective architectures such as {\inflow s} is that they can be trained using maximum likelihood. However, since the range of $f_\theta$ is a $d$-dimensional submanifold in $\R^D$, likelihoods of the samples not on this manifold are not defined. We circumvent this difficulty by splitting the training procedure into two phases---(i) mean squared error (MSE) training phase where we only optimize over the injective part ($g_\gamma$) of the network, and (ii) maximum likelihood (ML) training phase where we fit the parameters $\eta$ of the bijective part $h_\eta$ so as to maximize the likelihood of the preimage of training data through $g_\gamma$; this step  matches the density of $p_X$ to that of the ground truth $p_\Xi$.

The loss function that we minimize to find the parameters of $g_\gamma$ is given as
\begin{equation} \label{eq:lmse}
    \lmse(\gamma) = \dfrac{1}{N} \sum_{i=1}^N \| \xi^{(i)}- g_\gamma(g_\gamma^\dagger(\xi^{(i)}))\|_2^2
\end{equation}
where $\xi^{(i)}$-s are the training samples. We find that only a few epochs of training are sufficient to train $g_{\gamma}$. Note that $P_{g_{\gamma}}(x) := g_{\gamma}(g_{\gamma}^\dagger(x))$ is an idempotent projection operator on the range of $g_{\gamma}$. The low-dimensional range of $g_{\gamma}$ acts as a regularizer in the context of inverse problems. Injectivity implies that the range of $f_\theta$ is a true manifold unlike in the case of GANs where it may be an arbitrary low-dimensional structure~\citep{puthawala2020globally}. This allows us to define likelihoods as in \eqref{eq:cov_inj}.

After the MSE training phase, we have a manifold that near-interpolates the data samples. In the ML training phase, we match the density (or measure) on the manifold to $p_\Xi$ by maximizing the likelihood of the preimages of the training samples $\{ g_\gamma^\dag(\xi^{(i)})\}$ over $\eta$. This gives us the loss function for the ML training phase as
\begin{multline} \label{eq:loss_g}
    \lml(\eta)
    \\
    = \dfrac{1}{N}\sum_{i=1}^N \left(-\log p_Z(z^{(i)}) + \sum_{l=1}^{L} \log|\det J_{h_{\eta,l}}|\right),
\end{multline}
where $z^{(i)} = h_\eta^{-1}(g_\gamma^\dagger(\xi^{(i)}))$ and $J_{h_{\eta,l}}$ are evaluated at appropriate intermediate inputs. Such a stratified training strategy was proposed recently  by~\citet{brehmer2020flows}. They, however, concatenate regular bijective normalizing flows and pad zeros to the low-dimensional latent codes. This makes their method almost as compute intensive as regular flows.

\paragraph{Stability of layerwise inversions.} To minimize $\lmse$ \eqref{eq:lmse}, we need to calculate the left inverse $g_\gamma^\dag$ for points that do not lie in the range of $g_\gamma$. This entails computing the pseudoinverses of injective convolutional layers $\ell_w$. We study the stability of inversion for out-of-range points under the assumption that $y' = \ell_w(x) + \epsilon$, $\epsilon\sim\calN(0,\sigma_\epsilon^2 I)$. In particular, we are interested in estimating the inverse error $E_{\Inv}(y') = \|\ell_w^\dagger(y') - x\|_2^2$ and the re-projection error $E_{\Proj}(y') = \|\ell_w(\ell_w^\dagger(y')) - y'\|_2^2$. 

We show in Appendix \ref{app:derivations} that for both linear and $\Relu$ injective convolutions the average errors are
\[
\Exp{\epsilon}{E_{\Inv}(y)} \propto \sigma_\epsilon^2\sum_{i=1}^{c} \dfrac{1}{s_i(w)^2}, \qquad \Exp{\epsilon}{E_{\Proj}(y)} \propto \sigma_\epsilon^2,
\]
where $s_i(w)$-s are the singular values of $w$ and $c$ is the number of input channels in the forward direction.

The reconstruction error thus behaves gracefully in $\sigma_\epsilon$, but could blow up for poorly conditioned $w$. In order to stabilize inversions and training, we regularize the inverse via Tikhonov regularization. This changes the error terms from $\sum_{i=1}^{c} 1/s_i(w)^2$ to $\sum_{i=1}^{c} \frac{s_i(w)}{s_i(w)^2 + \lambda}$ which is upper bounded by $\frac{c}{2\sqrt{\lambda}}$, thus effectively stabilizing training. Here, $\lambda$ is the regularization parameter.

\section{Inference and uncertainty quantification with {\inflow}} \label{sec:inf_setup}

% \id{Perhaps open a new section here.}

% \id{Let us rather start with a Bayesian model, posteriors distribution, etc..., in the spirit UQ == Monte Carlo == Inverse Problems; then mention (perhaps inline) that this also models the usual $y = Ax$}\kk{Do you want to introduce Monte Carlo for sampling posterior? I am not sure what's the context. I have rewritten parts here to help address the rest. }

We consider reconstructing an object $x \in \R^D$ from measurements $y \in \R^n$. We assume that $x$ and $y$ are realizations of jointly distributed random vectors $X$, $Y$, with the joint distribution $p_{X, Y}(x, y)$.  In inference, we are mainly interested in characterizing the posterior $p_{X|Y}(x|y)$. We note that this setting generalizes point estimation of $x$ given $y$ common in inverse problems where the task is to recover $x$ from measurements $y = A x + \epsilon$. Here $\epsilon$ is additive noise and $A \in \R^{n \times D}$ is the forward operator. Examples of forward operators include the subsampled Fourier transform in magnetic resonance imaging (MRI) or a random matrix in compressed sensing. In many practical problems the number of measurements $n$ is much smaller than the number of unknowns to recover $D$. In such applications one often computes the maximum a posteriori (MAP) estimate  $x_\text{MAP} = \argmax_x p_{X|Y}(x|y)$; Bayes theorem yields
\begin{align} \label{eq:map_inv}
    x_{\text{MAP}} &= \argmin_x -\log p_{Y|X}(y|x) - \log p_{X}(x)\nonumber\\
    &= \argmin_x \tfrac{1}{2}\|y-Ax\|_2^2 - \sigma_\epsilon^2\log p_X(x),
\end{align}
where we assume that $\epsilon \sim \calN(0, \sigma_\epsilon^2 I)$.

\subsection{MAP estimation with \inflow~prior}

We now address two inference tasks where {\inflow s} are particularly effective. Recall that since $g_{\gamma}$ is injective one can build a fast projector $P_{g_{\gamma}}(x) = g_\gamma(g_\gamma^{\dagger}(x))$ on the range of $g_\gamma$, i.e., the range of our generator. 

% This in turn allows us to design a projected gradient descent method for MAP estimation with a {\inflow}~prior.
% Note that in order to compute the projection, earlier work involves a gradient-descent based inversion which can be extremely slow \textbf{[]}, or training a separate ``projection'' network. Here we use a fast, exact inverse. 

Beyond simply projecting on the range, injectivity and Bayes theorem enable us to maximize the likelihood of the reconstruction under the posterior induced by the \inflow~prior~\citep{whang2020compressed}. The injective flow (\iflow) algorithm described below in Algorithm \ref{alg:pgd} then alternates projections on the range with gradient steps on the data fidelity term and the prior density. We study two variants---\iflow~and \iflow-L that correspond to running Algorithm \ref{alg:pgd} without and with $-\log p_X$ terms. 
% \id{Should we emphasize that Algo 1 takes the log prior? by separating it out? I think we should and so the lambda becomes an explicit parameter as well.}

\begin{algorithm}[h!]
\SetAlgoLined
\SetKwFunction{grad}{GradientStep}
\SetKwFunction{proj}{$P_g$}
\KwIn{loss function $L, y, A, g_\gamma$ }
% \KwOut{$x = \argmin_{x\in\text{Range}(g)}  L(x)$ 
% \id{you used $\mathrm{Range}(f)$ for range before}}
\textbf{Parameter:} step size $\eta$ and $\lambda (\propto \sigma^2)$\;
\BlankLine
 $x^{[0]} = A^\dagger y$\;
 \For{$i\leftarrow 0$ \KwTo $T-1$}{
  $v \leftarrow$ \proj{$x^{[i]}$}\;
  $x^{[i+1]} \leftarrow$ \grad{$L(v)$}\;
 }
 $x^{[T]} \leftarrow P_g(x^{[T]})$\;
 \caption{\iflow}
 \label{alg:pgd}
\end{algorithm}

One caveat with computing $-\log p_X(x)$ is that it requires $\log|\det[J_{f_\theta}^\T J_{f_\theta}](f_{\theta}^\dagger(x))|$ according to \eqref{eq:cov_inj}. While we have layer-wise tractable Jacobians, $\log |\det J_{f_\theta}^\T J_{f_\theta}|$ cannot be split into layerwise $\log\det$ terms due to the change of dimension. Fortunately, the literature is abundant with efficient stochastic estimators. We describe one in Section \ref{sec:nll} that we use to compare and report likelihoods. In order to implement the \iflow-L, however, we propose a much faster scheme based on a bound.

We show in Appendix \ref{app:derivations} that
for an injective function $g:\R^d \mapsto \R^D$, where $g:=g_1\circ g_2 \ldots \circ g_K$, $\log|\det J_g^\T J_g| \le \sum_{i=1}^{K} \log|\det J_{g_i}^\T J_{g_i}|$.
% \label{prop:logdet}
% \end{prop}
Thus one gets an upper bound 
\begin{multline} \label{eq:bound_px}
   -\log p_X(x) \le -\log p_Z(f^{\dagger}(x))
   \\
   + \dfrac{1}{2}\sum_{k=1}^{K}\log |\det J_{g_{\gamma, k}}^\T J_{g_{\gamma, k}}| + \sum_{l=1}^{L} \log |\det J_{h_{\eta, l}}|,
\end{multline}
where the layer Jacobians are evaluated at the appropriate intermediate layer outputs. Since all our layers including the injective layers have $\log\det$ Jacobians readily available we use \eqref{eq:bound_px} as a proxy for $-\log p_X(x)$. Denoting the right-hand side of \eqref{eq:bound_px} by $R(x)$ yields the proposed iFlow-L algorithm (Algorithm \ref{alg:pgd}) for solving \eqref{eq:map_inv}. The objective function is 
\begin{equation} \label{eq:loss}
L(x) := \tfrac{1}{2}\|y-Ax\|^2_2 + \sigma^2 R(x).
\end{equation}
Note that when solving inverse problems we constrain the final solution $x$ to be in the range of $f$, that is, $x = f_\theta(z)$ for some $z \in \R^d$.

\subsection{Posterior modeling and uncertainty quantification}

The second application enabled by {\inflow s} is efficient uncertainty quantification for inverse problems in imaging.
% Many popular imaging algorithms are designed to produce a single reconstruction from the measurements. This is the case with minimization of regularized misfit functionals (such as the MAP estimator in the previous section) and with the end-to-end trained neural networks such as the U-Net. In ill-posed inverse problems, especially in the presence of noise, there could be many \textit{different} reconstructions that are equally likely under the posterior. One would ideally characterize the entire posterior distribution and then calculate uncertainty estimates for a given  reconstruction. 
We build on a method recently proposed by~\citet{sun2020deep} which computes a variational approximation to the posterior $p_{X|Y}(x|y)$ corresponding to the measurement $y$ and a ``classical'' regularizer. They train a normalizing flow which produces samples from the posterior, with the prior and the noise model given implicitly by the regularized misfit functional. 

The injectivity of the {\inflow}~generator $f_\theta$ and the assumption that the modeled data concentrates close to the range of $f_\theta$ allows us to write the posterior on $X$, $p_{X|Y}$, in terms of $p_{Z|Y}$, with $X = f_\theta(Z)$. That is,
\begin{equation}
    \label{eq:uq-posterior-z-or-x}
    p_{X|Y}(f_\theta(z) | y) = p_{Z|Y}(z | y) \cdot |\det J_{f_\theta}^T J_{f_\theta}|^{-1/2}.
\end{equation}
We can thus derive a computationally efficient version of the algorithm proposed by \citet{sun2020deep} by only training a low-dimensional flow. 

Instead of using {\inflow s} to simply reduce computational complexity, we showcase another interesting possibility: approximating the posterior with respect to the learned prior given by the {\inflow}. To do this we train another network $u_\upsilon$ which is a low-dimensional flow, so that the distribution of $  f_\theta \circ u_\upsilon(T)$ approximates the posterior $p_{X|Y}$ when $T$ is an iid Gaussian vector. The generative process for (approximate) samples from $p_{X|Y}$ is then
\[
    T \stackrel{u_\upsilon}{\longrightarrow} Z \underbrace{\stackrel{h_\eta}{\longrightarrow} Z' \stackrel{g_\gamma}{\longrightarrow}}_{f_\theta} X.
\]
% Assume $t \sim p_T$ and $f(u(t))$ generates samples from $p_{X|Y}$ where the random variable $X=x=f(u(t))$. Then instead of learning $p_{X|Y}$ one could learn $p_{Z|Y}$. 

% \[
% \begin{array}{ccccc}
%     Z & \stackrel{h}{\to} & Z' & \stackrel{g}{\to} & X \\
%     & \stackrel{u}{\searrow} & T & \stackrel{g}{\nearrow}
% \end{array}
% \]
We thus require that $u_\upsilon(T) \sim p_{Z|Y}$ with $T \sim \calN(0, I)$ and $X = f_\theta(Z)$. Letting $q_\upsilon$ be the distribution of $u_\upsilon(T)$, the parameters $\upsilon$ are adjusted by minimizing the KL divergence between $q_\upsilon$ and $p_{Z|Y}$,
\begin{multline} \label{eq:post_flow}
    \upsilon^* 
    = \argmin_{\upsilon}~ \mathrm{D}_{\text{KL}}\left(q_\upsilon \, \| \, p_{Z|Y} \right) \\
    = \argmin_{\upsilon}~ \Exp{Z \sim q_\upsilon} [-\log p_{Y|Z}(y | Z) - \log p_Z(Z) + \log q_\upsilon(Z)] \\
    = \argmin_{\upsilon} \Exp{T \sim \calN(0,I)} [-\log p_{Y|Z}(y | u_\upsilon(T)) - \log p_Z(u_{\upsilon}(T)) \\ 
    + \log p_T(T) - \log|\det J_{u_{\upsilon}}(T)|] . \\
\end{multline}

% Due to the increasing dimension of $f$, $\mathrm{dim}(z) \ll \mathrm{dim}(x)$ which makes learning \id{what do you mean by learning? training the network? that is certainly true. we should also give a statistical argument that it is much easier to do monte carlo integration over the much smaller domain of $z$ rather than that of $x$} the posterior $p(z|y)$ much cheaper. 

We revisit the inverse problem associated with $y = A x + \epsilon$ with $\epsilon \sim \calN(0, \sigma^2 I)$. In this setting we have 
\begin{equation} \label{eq:uqflow-invprob}
\begin{aligned}
    \upsilon^* &= \argmin_{\upsilon} \Exp{T \sim \calN(0,I)} \left[\tfrac{1}{2} \|y - A f_\theta(u_\upsilon(T))\|_2^2\right. \\ 
    &\quad-\left.\sigma^2\log p_Z(u_{\upsilon}(T))- \sigma^2\log|\det J_{u_\upsilon}(T)| \right].
\end{aligned}
\end{equation}
We evaluate \eqref{eq:uqflow-invprob} by drawing $k$ iid samples $\{ t_i\}_{i=1}^k$ from the base Gaussian, yielding the following loss to train $u_\upsilon$,
\begin{multline}
\calL(\upsilon) := \dfrac{1}{k}\sum_{i=1}^k(\|y-A f_\theta(u_\upsilon(t_k))\|_2^2
\\
- \sigma^2\log p_{Z}(u_\upsilon(t_k))
- \beta\sigma^2\log|\det J_{u_\upsilon}(t_k)|),
\end{multline}
where we added $\beta$ as a hyper-parameter to control the diversity of samples we generate from the posterior~\citep{sun2020deep}. 

\subsection{Estimating log-likelihoods} \label{sec:nll}

The training of {\inflow s} only requires the log det of the Jacobian of $h_\eta$. Some applications call for the log det of the Jacobian of the full network, typically evaluated a small number of times. Here, we provide a stochastic estimate via the truncation of a Neumann series.

As $J_{f_{\theta}}^\T J_{f_{\theta}}$ is a square matrix, we find that 
\begin{align*}
\log |\det J_{f_{\theta}}^\T J_{f_{\theta}}| &= \Tr(\log J_{f_{\theta}}^\T J_{f_{\theta}})  \\[0.25cm]
&=\Tr\left(\log \dfrac{1}{\alpha}(I-(I-\alpha J_{f_{\theta}}^\T J_{f_{\theta}}))\right)\\
&=-\Tr\left(\sum_{k=1}^{\infty} \dfrac{(I-\alpha J_{f_{\theta}}^\T J_{f_{\theta}})^k}{k}\right) - d\log\alpha \\
&\approx -\Exp{v}{\sum_{k=1}^{n} \dfrac{1}{k} v^\T(I-\alpha J_f^\T J_f)^k v} - d\log\alpha
\end{align*}
where we choose $\alpha$ such that the maximal singular value of $I-\alpha J_{f_{\theta}}^\T J_{f_{\theta}}$ is about $0.1$. This ensures that the series converges fast and we can truncate the expansion to about $10$ terms. We estimate the largest singular value of $J_{f_{\theta}}^\T J_{f_{\theta}}$ using power iteration. In the last step we use the Hutchinson trace estimator~\citep{hutchinson1989stochastic} to evaluate the trace. Here, $v$s are sampled from $\calN(0,I)$. The terms of the power series can be efficiently implemented by vector-Jacobian and Jacobian-vector products using automatic differentiation as described in Algorithm \ref{alg:stochastic_logdet_estimator}~\cite{chen2019residual}.
% where $n \sim p(N)$ and $v \sim \calN(0,I)$. The last step is the ``Russian roulette'' estimator~\cite{chen2019residual,kahn1955use}.

% The gradient of the $\log|\det J_f^\T J_f|$ w.r.t theta may be required

\begin{algorithm} 
\SetAlgoLined
\SetKwFunction{vjp}{vjp}
\SetKwFunction{jvp}{jvp}
\SetKwFunction{maxsing}{MaxSingularValue}
\KwIn{$f, n$ }
\KwOut{$\log|\det J_f^\T J_f|$}

\BlankLine
$\log\det = 0$\\
$\beta = 0.9 \left(\maxsing(J_f)\right)^{-1}$\;
Draw $v$ from $\calN(0,I)$\;
$w^\T = v^\T$\;
 \For{k=1 \KwTo n}{
    $u_1^\T = \jvp(w)$\;
    $u_2^\T = \vjp(u_1)$\;
    $w = w - \beta u_2$\;
    % $p = \Pr(N\ge k)$\;
    $\log\det \mathrel{-}= \dfrac{w^\T v}{k}$\;
 }
 $\log\det \mathrel{-}= d\log\beta$
 
\caption{Stochastic $\log\det$ Jacobian estimator}
\label{alg:stochastic_logdet_estimator}
\end{algorithm}

\section{Computational experiments with imaging problems} \label{sec:exp}

We begin by evaluating the generative performance of {\inflow s}. Next, we test {\inflow s} on 
two inference tasks in imaging: maximum a posteriori estimation and uncertainty quantification. .

\subsection{Generative modeling}

We train {~\inflow s} on the MNIST~\citep{lecun1998gradient}, CIFAR10~\citep{krizhevsky2009learning},  CelebA~\citep{liu2015deep}
% , LSUN Churches~\citep{yu2015lsun} 
and Chest X-ray~\citep{wang2017chestx} datasets with image sizes  $32\times32\times 1$, $32\times32\times 3$, $64\times64\times3$
% , $64\times64\times3$ 
and $128\times128\times1$ respectively.

We find that our networks train much faster than invertible flows and their recent injective generalizations~\citep{brehmer2020flows}. As a point of comparison, training the models of ~\citet{brehmer2020flows} takes over 10 days on the CelebA dataset. The corresponding \inflow~trains in 38 hours while yielding better samples in terms of the Fr\'echet inception distance (FID)~\citep{heusel2017gans} (see Table \ref{tab:comp_fid}). \footnote{Our FID scores are reported at sampling temperature $T=1$, that is, we use the same prior distribution statistics for training and sampling. We show the variation of the FID metric with the temperature in Figure \ref{fig:celeba_fid} in Appendix \ref{app:samples}}

Since the range of a {~\inflow } is a manifold, a relevant metric is the reconstruction error, $\frac{\|\xi-f_\theta(f_\theta^\dagger(\xi))\|}{\|\xi\|}$, which we report for $\xi$s in the test set in Table \ref{tab:comp_resource}. We share generated samples and reconstructions on test sets from trained {\inflow s} in Figures \ref{fig:mnist_linear_conv}, \ref{fig:celeba_linear_conv}, \ref{fig:chest_samples}
% ,\ref{fig:church_samples}
and \ref{fig:cifar10_samples}  in Appendix \ref{app:samples}. 

We note that the variants with the linear and ReLU 1 $\times$ 1 convolutions perform similarly (see Figures \ref{fig:mnist_relu_conv},\ref{fig:mnist_linear_conv}, \ref{fig:celeba_relu_conv}, \ref{fig:celeba_linear_conv}); hence, for the subsequent datasets and experiments we only report results with the linear variant.

\begin{table}[]
\centering
\caption{FID scores on $8$-bit $64\times64$ celebA dataset.}
\label{tab:comp_fid}
% \resizebox{0.47\textwidth}{!}
{%
\begin{tabular}{@{}lc@{}}
\toprule
Model & FID \\ \midrule
\cite{kumar2020regularized}     &  40.23 \\
\cite{brehmer2020flows} & 37.4 \\
\inflow~(Ours) & \textbf{34.3}  \\\bottomrule
\end{tabular}%
}
\end{table}

% \kk{Ivan and Maarten: Please check if these can be converted to bits per dimension.} 
The negative log-likelihood values estimated for trained \inflow~ models using Algorithm \ref{alg:stochastic_logdet_estimator} on the $[-1,1]$ normalized MNIST and CelebA dataset are $114.82\pm 5.8$ and $294 \pm 7.4$ nats respectively.

% Our parameter count on the same dataset is about $16$M. Hence, just having fewer parameters cannot be responsible for the speed-up. We believe that since natural image data lies on low-dimensional structures in a high-dimensional space a bijective map in the high-dimensional space forces the Jacobian to become close to low rank around data samples. A poorly conditioned Jacobian can cause gradients to become unstable requiring careful clipping and regularization to be imposed while training~\citep{kingma2018glow,brehmer2020flows}. Since {\inflow s} model a true low-dimensional manifold as their range, they suffer less from such issues. 

\begin{table}[]
\centering
\caption{Training times in hours for \inflow: all models were trained on a single V100 GPU}
\label{tab:comp_resource}
\resizebox{0.47\textwidth}{!}{%
\begin{tabular}{@{}lccc@{}}
\toprule
        & \begin{tabular}[c]{@{}c@{}}Training \\ time (hours)\end{tabular} & $\dfrac{\|x-f_\theta(f_\theta^\dagger(x)\|}{\|x\|}$ & \begin{tabular}[c]{@{}c@{}}Trainable \\ params\end{tabular} \\ \midrule
MNIST     & $11$      & $0.04$                                                             & $9$M \\

CIFAR10 & $11$      & $0.22$                                                               & $9$M    \\

CelebA & $38$       & $0.15$                                                               & $16$M \\

% LSUN Church & $27$      & $0.37$                                                               & $11$M \\
Chest X-ray & $25$      & $0.13$                                                               & $11$M \\\bottomrule
\end{tabular}%
}
\end{table}

\subsection{MAP estimation}

We test {\inflow s} on image reconstruction from compressive measurements. We work with four different forward operators / corruption models: (i) \textbf{RandGauss (m)}: we sample an entrywise iid Gaussian matrix $A\in\R^{n\times D}$, where $n=250$ and $D$ is the dimension of the vectorized image; (ii) \textbf{RandMask (p)}: we mask pixels (that is, replace a pixel with zero) with probability $p=0.15$; (iii) \textbf{Super-resolution (x4)}: we downsample the image by a factor of $4$ along each dimension; and (iv) \textbf{Mask (s)}: we mask (replace with zero) an $s \times s$-size portion of the image.

\begin{figure}
    \centering
    \includegraphics[width=0.47\textwidth]{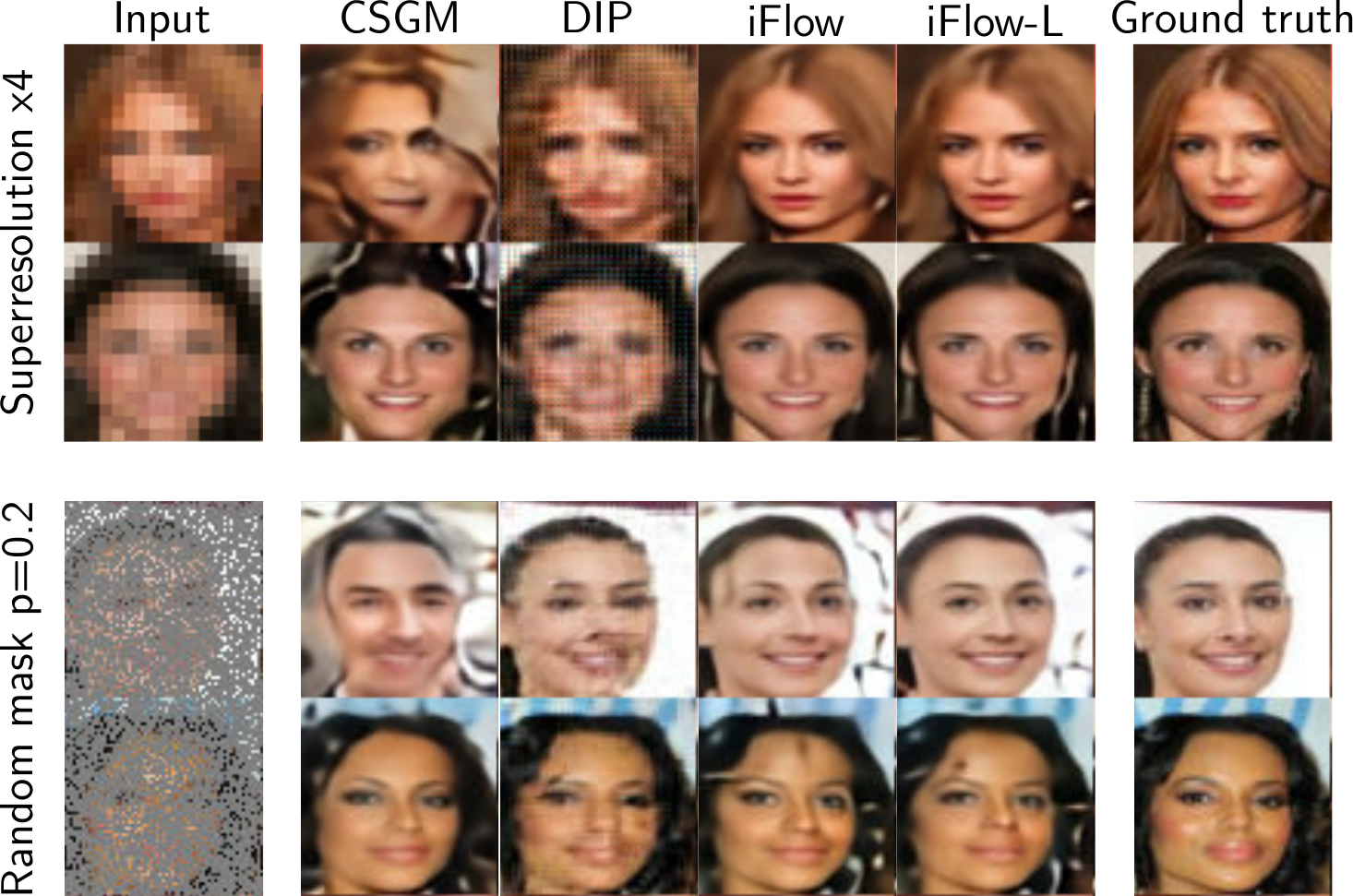}
    \caption{Comparison of various reconstruction schemes. The \iflow-L and \iflow~ methods refer to Algorithm \ref{alg:pgd} respectively with and without the likelihood term.}
    \label{fig:inv_probs_celeba}
\end{figure}

Since \inflow s have a readily available inverse we focus on the benefits this brings in imaging. Specifically, we use Algorithm \ref{alg:pgd} to compute an estimate using a trained \inflow~prior. We test the algorithm on the MNIST and CelebA datasets and use the same \inflow~prior for all problems. We compare our approach to two deep learning baselines---compressed sensing with generative models (CSGM) \citep{bora2017compressed} and deep image prior (DIP) \citep{ulyanov2018deep}. CSGM solves $\hat{x} = f(\argmin_z \|y - Af(z)\|_2^2)$ while DIP solves $\hat{x} = f_{\theta}(\argmin_\theta \|y - Af_\theta(z)\|_2^2)$ given a randomly chosen fixed $z$ and regularized by early stopping. Figure \ref{fig:inv_probs_celeba} compares all methods for the superresolution and random masking problems on the CelebA dataset while Table \ref{tab:compressed_sensing} gives a comprehensive evaluation for all inverse problems. 

We also perform an ablation study to assess the influence of including the prior likelihood as opposed to simply doing a gradient descent with manifold projections~\citep{raj2019gan}. The latter corresponds to setting $\lambda = 0$ in Algorithm \ref{alg:pgd}.
Table \ref{tab:compressed_sensing} clearly shows that  accounting for the prior density and not only support---that is, computing the MAP estimate---performs better in almost all settings.

\begin{table}[]
\renewcommand{\arraystretch}{1.2} 
\centering
\caption{Performance on inverse problems measured in reconstruction SNR (dB)}
\label{tab:compressed_sensing}
\resizebox{0.47\textwidth}{!}{%
\begin{tabular}{@{}llcccc@{}}

\toprule
                                       & \multicolumn{1}{c}{Dataset} & CSGM  & DIP   & \iflow & \iflow-L \\ \midrule
\multirow{2}{*}{\textit{RandGauss} $(m=250)$}     & MNIST              & 11.32 & 12.72 & 21.34 & \textbf{21.81}   \\\vspace{1mm}
                                       & CelebA             & 8.98     & \textbf{11.25}     &    8.90   &    8.91     \\
\multirow{2}{*}{\textit{RandMask} $(p=0.15)$}     & MNIST              & 3.85  & 4.94  & 4.76  & \textbf{10.10}   \\\vspace{1mm}
                                       & CelebA             & 12.63 & \textbf{17.26} & 13.89 & 14.43   \\
\multirow{2}{*}{\textit{Super-resolution} ($\times 4$)} & MNIST              & 5.943 & 1.0   & 9.851 & \textbf{12.75}   \\\vspace{1mm}
                                       & CelebA             & 11.08 & 14.12 & 17.36 & \textbf{20.07}   \\
\multirow{2}{*}{\textit{Mask} ($s=15$ px)}        & MNIST              &    3.34    &  4.38     &  3.90    &  \textbf{9.54}        \\
                                       & CelebA             & 13.42 & 21.31 & 21.74 & \textbf{21.79}   \\ \midrule
\textit{Limited-view CT}         & Chest              &    11.58    &  13.76     &  20.93    &  \textbf{21.23}        \\ \bottomrule
\end{tabular}%
}
\end{table}

We mention that we attempted to compare with a method involving projections proposed by \citet{shah2018solving} but found it to be  $50-100\times$ slower than \iflow. It was thus infeasible to finalize this comparison. On average we found that DIP converged the fastest followed by our method followed which was about $2\times$ slower. Finally, while each iteration of CSGM was as fast as each of DIP, CSGM requires several restarts which made the method about $4x$ slower than ours. We report the best results from CSGM with 10 restarts.

Note that the baselines \citep{bora2017compressed,ulyanov2018deep,shah2018solving} were developed without injectivity as a constraint. As a result they typically use off-the-shelf GAN architectures inspired by \citep{radford2015unsupervised}, but they are by design agnostic to architectural details. Therefore, in order to keep the comparisons fair, we use the same generative model $f_\theta$ for all methods. This allows us to test the importance of tractable inverses and likelihoods for the design of image reconstruction algorithms based on generative priors. 

\begin{figure*}[ht] 
\includegraphics[width=\textwidth]{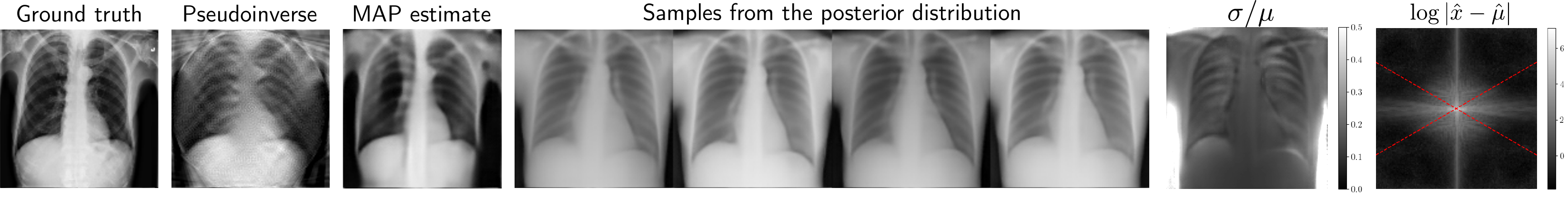}
\caption{Uncertainty quantification for limited view CT.}
\label{fig:uq}
\end{figure*}

\subsection{Posterior modeling and uncertainty quantification}

Next, we use \inflow~ priors for uncertainty quantification in computed tomography. We work with a chest $X$-ray dataset and use the limited-angle CT operator as the forward operator, $A$. We choose a sparse set of $n_\text{angles}=30$ view angles from $30^\circ$ to $150^\circ$, with a $60^\circ$ missing cone. We add $30$dB noise to the measurements. The resulting inverse problem is severely ill-posed and solving it requires regularization. (Note that Table~\ref{tab:compressed_sensing} includes the performance of Algorithm \ref{alg:pgd} on this problem.)

Here we provide a pixel-wise uncertainty estimate of the form $\Exp{X \sim p_{X|Y=y}}{|X - \langle X \rangle|^p}$, with $p = 1, 2$,  $| \cdot |$ the pixel-wise absolute value, and $\langle X \rangle$ the posterior mean. 
% \md{can we illustrate MCMC based on the latent space} 
In Figure~\ref{fig:uq}, we show the MAP estimate obtained from the \iflow-L algorithm (Algorithm \ref{alg:pgd}). We also show the Fourier spectrum of the mean absolute deviation calculated in the Fourier domain where the mean was calculated over the Fourier transform of all samples from the posterior. We observe a cone of increased uncertainty in the Fourier spectrum that corresponds to the missing angles in the limited-view CT operator. Furthermore, we observe a thick vertical bright line that corresponds to uncertainty in predicting the location of the ribs (which have a strong horizontal periodic component) as shown in the middle plot of Figure~\ref{fig:uq}. 

Reassuringly, both the spatial- and the frequency-domain representations of uncertainty correlate well with our intuitive expectations for this problem. Positions of the ribs in space and the missing cone in the spectrum exhibit higher uncertainty.

% \begin{figure} \label{fig:recon}
%     \centering
%     \includegraphics[width=0.45\textwidth]{graphics/recon.pdf}
%     \caption{Reconstruction of CelebA samples}
%     \label{fig:cs}
% \end{figure}

% [redo, talk about how having the inverse helps] Due to a readily availabe inverse in our proposed injective flows, going from $x$-space to $z$-space is easy and helps improve our performance over baselines both qualitatively and quantitatively in the SNR metric. 

% Recently a method to quantify uncertainty in inverse problems with flow priors was proposed~\cite{sun2020deep}. The algorithm relies on initializing the optimization multiple times with different latent vector codes $z$ and using the different results as proxy to sampling the posterior. Using a similar strategy we run Algorithm \ref{alg:pgd} for multiple instances with different random noise vectors added to the initial guess, $x^{(0)} = A^\dagger y$ and check their standard deviation. 

% \begin{figure} \label{fig:cs_results}
%     \centering
%     \includegraphics[width=0.45\textwidth]{graphics/cs.pdf}
%     \caption{Compressed sensing results. Ground truth for celebA can be found in Figure \ref{fig:recon}}
%     \label{fig:cs}
% \end{figure}

\section{Related work}
\label{sec:related-work}

Normalizing flows have been introduced in~\citep{dinh2014nice}. The key to their success are invertible coupling layers with triangular Jacobians. Different variants of the coupling layer along with multiscale architectures~\citep{dinh2016density, kingma2018glow, grathwohl2018ffjord} have considerably improved performance of normalizing flows. Glow~\citep{kingma2018glow} uses invertible $1\times 1$ convolutions to improve expressivity, producing better samples than NICE and Real-NVP. Alas, training a Glow model is extremely compute intensive---1 week on 40 GPUs for the 5-bit $256\times 256$ CelebA dataset. A crucial drawback of the mentioned models is that they are bijective so the dimension of the latent and data spaces coincide. This results in a large number of parameters and slow training: since the ground data lies close to low-dimensional subset of $\R^D$, training should encourage the model to become ``almost non-invertible'' which makes the optimization more difficult.

\citet{kumar2020regularized} propose approximate injective flows by using spectral regularization in auto-encoders. However they lack access to likelihoods. Further, their training strategy is only a proxy for injectivity. Very recently, \cite{brehmer2020flows} proposed injective flows to learn a data distribution on a manifold very similar to our work, including a two-stage training scheme we use. However, they use regular normalizing flow architectures with zero padding in the latent space which results in architectures that are very expensive to train. \citet{cunningham2020normalizing} build injective flows by adding noise to the range; this requires stochastic inversion whereas ours is deterministic.

In a parallel development, autoregressive flows were shown to have favorable expressivity compared to normalizing flows. We refer to ~\citet{papamakarios2017masked,kingma1606improving,pmlr-v48-oord16} and the references therein for a more extensive account.

% Inverse autoregressive flows, masked autoregressive flows, real nvp, ffjord, glow, PixelCNN, pix2pix

% Ben and Kumar, relaxed injective flows, NYU paper, simultaneous flow and manifold estimation, normalizing flows across dimensions, wavelet flows

\section{Discussion and conclusion} \label{sec:discussion}

We proposed {\inflow s}---a flow-based generative model that is injective by construction.
{\inflow s} alleviate the main drawback of invertible normalizing flows which is that they are very expensive to train. We showed that {\inflow s} are competitive in terms of generative modeling performance and that the fast inverse on the range markedly improves reconstructions in ill-posed inverse problems. We also showed how to use {\inflow s} to model posteriors and perform uncertainty quantification directly in the low-dimensional latent space.
% Similarly, if we have a low-dimensional prior, then Monte Carlo sampling of the posterior should be performed in the low-dimensional space.
Currently our reconstructions on data lack high frequency features. This is common in normalizing flow models~\citep{dinh2016density}. Strategies such as adding the adversarial loss in the MSE phase of training may help alleviate this drawback. Furthermore, using a richer class of coupling layers may help--- \citet{durkan2019neural} show that flows based on rational quadratic splines are more expressive. Integrating such layers also holds promise for improving the expressivity of \inflow s.

Our work combines a number of basic ideas in an intuitive way that yields  gains in efficiency and accuracy. Additionally, recent results on universality of globally injective neural networks \citep{puthawala2020globally} and universality of flows \citep{teshima2020coupling} suggest that {\inflow s}~are universal approximators of measures concentrated on Lipschitz manifolds; a rigorous proof is left to future work.

% : Given a source probability measure supported in some compact subset of $\mathbb{R}^d$ which is absolutely continuous with respect to the Lebesgue measure, and a target probability measure which is supported on a $d$-Hausdorff dimension manifold embedded in $\mathbb{R}^D$ ($D \gg d$) which is continuous with respect to the Hausdorff measure on that manifold, we can always find a {\inflow}~to map the source measure to the target measure. The proof will be given in a forthcoming paper \citep{puthawalaUQ}. \md{please add a reference to our next paper}

\section*{Acknowledgements}
MVdH gratefully acknowledges support from the Department of Energy under grant DE-SC0020345, the Simons Foundation under the MATH + X program, and the corporate members of the Geo-Mathematical Imaging Group at Rice University. ID and AK were supported by the European Research Council Starting Grant 852821---SWING.

\bibliography{example_paper}
% \bibliographystyle{icml2021}

%%%%%%%%%%%%%%%%%%%%%%%%%%%%%%%%%%%%%%%%%%%%%%%%%%%%%%%%%%%%%%%%%%%%%%%%%%%%%%%
%%%%%%%%%%%%%%%%%%%%%%%%%%%%%%%%%%%%%%%%%%%%%%%%%%%%%%%%%%%%%%%%%%%%%%%%%%%%%%%
% DELETE THIS PART. DO NOT PLACE CONTENT AFTER THE REFERENCES!
%%%%%%%%%%%%%%%%%%%%%%%%%%%%%%%%%%%%%%%%%%%%%%%%%%%%%%%%%%%%%%%%%%%%%%%%%%%%%%%
%%%%%%%%%%%%%%%%%%%%%%%%%%%%%%%%%%%%%%%%%%%%%%%%%%%%%%%%%%%%%%%%%%%%%%%%%%%%%%%
\clearpage
\appendix

\section{Network architecture and training details} \label{app:net_arch}

We describe the injective portion of our network architecture that was used to train a CelebA dataset in Figure \ref{fig:network_detail}. The bijective revnet block has 3 bijective revnet steps in each block while the injective revnet block has just one injective revnet step which is explained in details in Section \ref{sec: injective_flow}.  The bijective part of our network is not shown in Figure \ref{fig:network_detail} but it has 32 bijective revenet steps.

For the scale and bias terms of the coupling layer we used the U-Net architecture with 2 downsampling blocks and 2 corresponding upsampling blocks. Each resolution change is preceded by 2 convolution layers with $32$ and $64$ output channels. We choose the latent space dimension as $64$ for MNIST, $256$ for Chest X-ray dataset and $192$ for all other datasets. We normalize the data to lie in $[-1, 1]$.

The number of training samples for CelebA, Chest X-ray, MNIST and CIFAR10 are 80000, 80000, 60000, and 50000 respectively. We trained all models for about 300 epochs with a batch size of 64.

All models are trained with Adam optimizer \citep{kingma2014adam}  with learning rate $10^{-4}$. $\gamma = 10^{-6}$ was used as the Tikhonov regularizer parameter for computing  pseudoinverse of injective convolutional layers.

\begin{figure}
    \centering
    \includegraphics[width=0.31\textwidth]{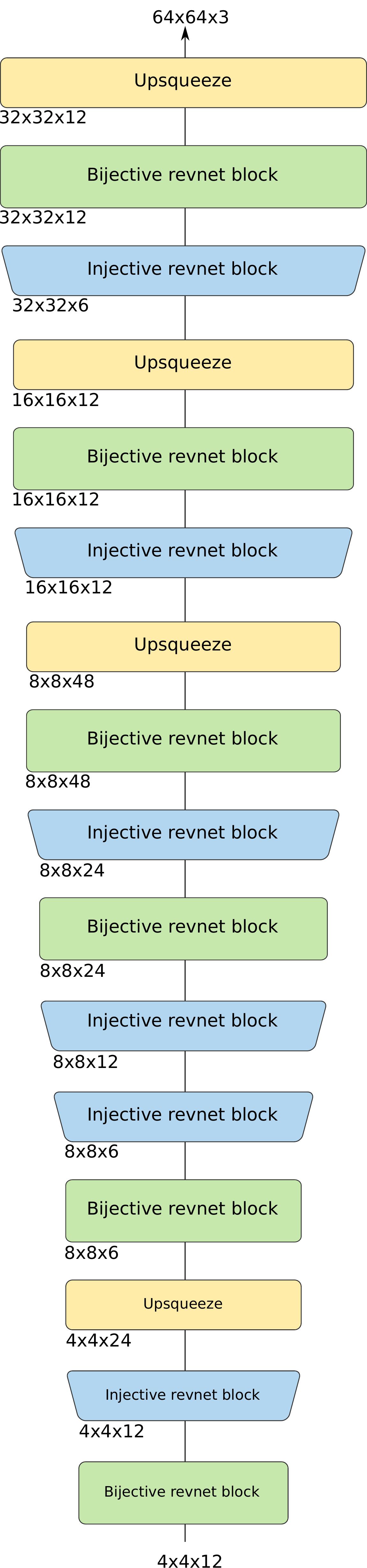}
    \caption{CelebA architecture for the injective portion $g$ of \inflow. The input size to each layer is written below it.}
    \label{fig:network_detail}
\end{figure}

\section{Derivations of error and likelihood bounds} \label{app:derivations}

\subsection{Bounding log-likelihood for injective functions}
\begin{prop}
For an injective function $f=f_1\circ f_2 \circ \ldots f_k(z)$ that maps $z\in\R^d$ to $x\in\R^D$,
\[
\log|\det J_f^\T J_f| \le \sum_{i=1}^{K} \log|\det J_{f_k}^\T J_{f_k}|
\]
\end{prop}
\begin{proof}
We demonstrate the claim for 3 layers; the general statement follows by induction. Consider $x = f(z) = f_1 \circ f_2 \circ f_3 (z)$, where $x\in \R^D$ and $z\in \R^d,\ d<D$. Assume that $f_1:\R^D \mapsto \R^D,\ f_3:\R^d \mapsto \R^d$ are bijective and $f_2:\R^d \mapsto \R^D$ is injective. Then 
\[
J_f = \underbrace{\dfrac{\partial f_1}{\partial f_2}}_{J_1}\underbrace{\dfrac{\partial f_2}{\partial f_3}}_{J_2}\underbrace{\dfrac{\partial f_3}{\partial z}}_{J_3}
\]
and we have
\begin{align} \label{part1}
\log\det|J_f^\T J_f| &= \log\det|J_3^\T J_2^\T J_1^\T  J_1 J_2 J_3 |\nonumber\\
&=2\log\det|J_3| + \log\det|J_2^\T J_1^\T J_1 J_2  |.
\end{align}
Let now $J_1 = U_1 \Sigma_1 V_1^\T$ and $J_2 = U_2 \Sigma_2 V_2^\T$. We can compute as
\begin{align} \label{part2}
  \log|\det J_2^\T J_1^\T J_1 J_2 | &= \log|\det V_2 \Sigma_2 U_2^\T V_1 \Sigma_1 U_1^\T U_1 \Sigma_1 V_1^\T U_2 \Sigma_2 V_2^\T   |  \nonumber\\
  &= \log |V_2 \Sigma_2 U_2^\T V_1 \Sigma_1^2 V_1^\T U_2 \Sigma_2 V_2^\T|\nonumber\\
  &= 2\log |\det \Sigma_2| + \log |\det V_2^\T U_1 \Sigma_1^2 U_1^\T V_2 |\nonumber\\
  &\le 2\log |\det \Sigma_2|  + 2\log |\det \Sigma_1| \nonumber\\
  &= \log |\det J_2^{\T} J_2|  + \log |\det J_1^{\T} J_1| 
\end{align}
where we used that $\Pi_{i=1}^{n} \lambda_i(UHU^\T) \le \Pi_{i=1}^{n}\lambda_i(H)$ for any symmetric matrix $H$ and unitary matrix $U$ (\citet{horn1950singular}). Here $\lambda_i(M)$ is the $i$th eigenvalue of matrix $M$.

Substituting \eqref{part2} in \eqref{part1} we obtain,
\[
\log|\det J_f^\T J_f| \le \sum_{i=1}^{3} \log|\det J_{f_k}^\T J_{f_k}| ,
\]
which establishes the claim.
\end{proof}

\subsection{Measuring error due to deviations from range}
\begin{prop}
Consider $y' = y + \epsilon$, $\epsilon\sim\calN(0,\sigma_\epsilon^2 I)$, $y=\ell_w(x)$ and let $E_{\Inv}(y') := \|\ell_w^\dagger(y') - x\|_2^2$ and the re-projection error $E_{\Proj}(y') := \|\ell_w(\ell_w^\dagger(y')) - y'\|_2^2$. Then for both $\Relu$ and linear variants of $\ell_w$ we have
\[
\Exp{\epsilon}{E_{\Inv}(y')} \propto \sigma_\epsilon^2\sum_{i=1}^{c} \dfrac{1}{s_i(w)^2}, \qquad \Exp{\epsilon}{E_{\Proj}(y')} \propto \sigma_\epsilon^2,
\]
where $s_i(w)$'s are the singular values of $w$ and $c$ is the number of input channels in the forward direction.
\end{prop}
\begin{proof}

Consider $y' = y + \epsilon$, where $y = \ell_w(x)$ and $\epsilon \sim \mathcal{N}(0, \sigma^2_{\epsilon}I_{2n})$. We consider a vectorized $x$ and write the $1\times 1$ convolution as a matrix-vector product, $Wx$ say. For a $\Relu$ injective convolution one could write the inverse as
\begin{equation}
x' = W^\dagger\begin{bmatrix} I_n & -I_n \end{bmatrix}y'.
\end{equation}
We calculate $\Exp{\epsilon}{\|x' - x\|^2_2}$.
Let $M:=\begin{bmatrix} I_n & -I_n \end{bmatrix}$ and $B := W^\dagger $, then

\begin{align*}
x' &= BM (y + \epsilon)\\
x'-x &=  B M\epsilon ,
\end{align*}
whence
\begin{align*}
\|x' - x\|^2_2 &= (B M\epsilon)^\T B M\epsilon\\
\|x' - x\|^2_2 &= \Tr\left( B M\epsilon (B M\epsilon)^\T\right)\\
\|x' - x\|^2_2 &= \Tr\left( B M\epsilon\epsilon^\T M^\T B^\T\right)\\
\|x' - x\|^2_2 &= \Tr\left( M^\T B^\T B M\epsilon\epsilon^\T \right)
\end{align*}
so that
\begin{align*}
\Exp{\epsilon}{\|x' - x\|^2_2} &= \Exp{\epsilon}{\Tr\left( M^\T B^\T B M\epsilon\epsilon^\T \right)}\\
\Exp{\epsilon}{\|x' - x\|^2_2} &= \Tr\left( M^\T B^\T B M \right)\sigma^2_{\epsilon}\\
\Exp{\epsilon}{\|x' - x\|^2_2} &= 2\Tr\left( B^\T B\right)\sigma^2_{\epsilon}\\[-0.1cm]
\Exp{\epsilon}{\|x' - x\|^2_2} &= 2\sum_{i=1}^{c}s_i(w)^{-2}\sigma^2_{\epsilon} .\qquad\qedsymbol
\end{align*}

Similarly for a linear layer the inverse is given as $x'=By'$. Therefore,
\begin{align*}
x' &= B (y + \epsilon)\\
x'-x &=  B \epsilon
\end{align*}
whence
\begin{align*}
\|x' - x\|^2_2 &= (B \epsilon)^\T B \epsilon\\
\|x' - x\|^2_2 &= \Tr\left( B \epsilon (B \epsilon)^\T\right)\\
\|x' - x\|^2_2 &= \Tr\left( B \epsilon\epsilon^\T B^\T\right)\\
\|x' - x\|^2_2 &= \Tr\left( B^\T B \epsilon\epsilon^\T \right)
\end{align*}
so that
\begin{align*}
\Exp{\epsilon}{\|x' - x\|^2_2} &= \sum_{i=1}^{c}s_i(w)^{-2}\sigma^2_{\epsilon}.\qquad\qedsymbol
\end{align*}

The re-projection error for a $\Relu$ layer is given as
\begin{align*}
    E_\text{Proj}(y') 
    &= \left\| \Relu \left(\wmw x'\right) - y \right\|^2 \\
    &= \left\| \Relu \left(\wmw x'\right) - \Relu\left(\wmw x \right) - \epsilon \right\|^2 \\
    & \leq \left\| \wmw x'  - \wmw x \right\|^2 + \|\epsilon\|^2\\
    &= \left\| \wmw (x + B M \epsilon) - \wmw x \right \|^2 + \|\epsilon\|^2\\
    &= \left\| \wmw B M \epsilon \right \|^2 \\[0.45cm]
    & \leq  (2\| W W^\dagger\|^2 + 1) \|\epsilon \|^2\\[0.65cm]
    &= (2c+1)\|\epsilon \|^2.\qquad\qedsymbol
\end{align*}

Similarly, for a linear layer we have
\begin{align*}
    E_\text{Proj}(y') 
    &= \left\| Wx' - Wx - \epsilon \right\|^2 \\
    &= \left\| WW^\dagger\epsilon - \epsilon \right\|^2 \\
    &=  (c+1)\left\|\epsilon \right\|^2 .
\end{align*}
\end{proof}

\subsection{log-determinants of Jacobians for ReLU injective convolutions}

We vectorize $x$ and, again, write the $1 \times 1$ convolution as a matrix-vector product $W x$. Then, for a $\Relu$ $1 \times 1$ convolution, we have
\[
   y = \relu{\pmw{W}} x .
\]
This could be trivially rewritten as $y = W'x$, where the rows of $W'$ are $w'_i = w_i$ if $\langle w_i, x \rangle > 0$ and $w'_i = -w_i$ otherwise. We note that changing the row signs does not change $|\det W|$. Hence, for such a $\Relu$ injective convolutional layer, $\ell_w$  $\log|\det J_{\ell_w}^T J_{\ell_w}|= \sum_{i=1}^c s_i^2(w)$, where $s_i(w)$'s are the singular values of $w$, where $w$ is the $1\times 1$ kernel corresponding to the convolution matrix $W$.

\section{Samples}
\label{app:samples}

In Figures \ref{fig:mnist_relu_conv}, \ref{fig:mnist_linear_conv} and Figures \ref{fig:celeba_relu_conv}, \ref{fig:celeba_linear_conv} we compare the performance of {\inflow s} trained with $\Relu$ and linear injective convolutions on the MNIST and $64\times64$ CelebA datsets. Both variants offer similar  performance hence we choose to use linear convolutions for the rest of our results regarding inverse problems and uncertainty quantification. In Figures \ref{fig:cifar10_samples}
% , \ref{fig:church_samples}
and \ref{fig:chest_samples} we show generated samples from \inflow~ and a few reconstructions of original samples, $x$ given as $f(f^\dagger(x))$ on the CIFAR10 and Chest X-ray datasets respectively. For the 
% LSUN churches and 
CIFAR10 dataset, we do see a low frequency bias in the generated samples. We hope to rectify this as per our discussions in Section \ref{sec:discussion}. For other datasets the low-frequency bias seems to be less of a problem. In fact, on these datasets {\inflow s}  outperform previous injective variants of flows~\citep{brehmer2020flows,kumar2020regularized}.

The temperature of sampling has a significant effect on the FID scores as shown in Figure \ref{fig:celeba_fid}. While samples in Figures \ref{fig:celeba_relu_conv}, \ref{fig:celeba_linear_conv} are for $T=1$ we share some samples in Figure \ref{fig:celeba_linear_0.85} for $T=0.85$.
\begin{figure}
    \centering
    \includegraphics[width=0.47\textwidth]{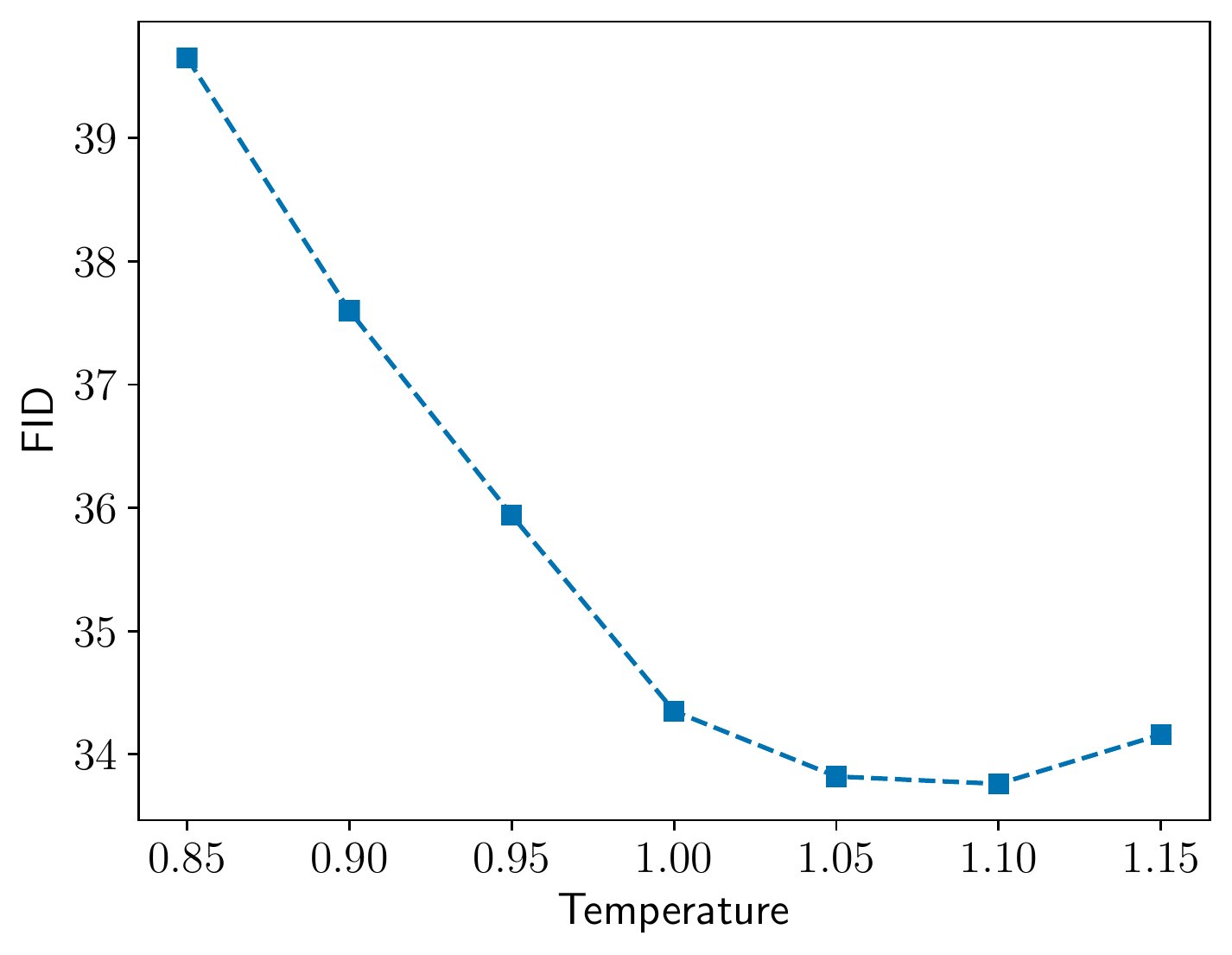}
    \caption{FID score of \inflow~with sampling temperature.}
    \label{fig:celeba_fid}
\end{figure}

\begin{figure*}
    \centering
    \begin{subfigure}{0.9\textwidth}
    \includegraphics[width=\textwidth]{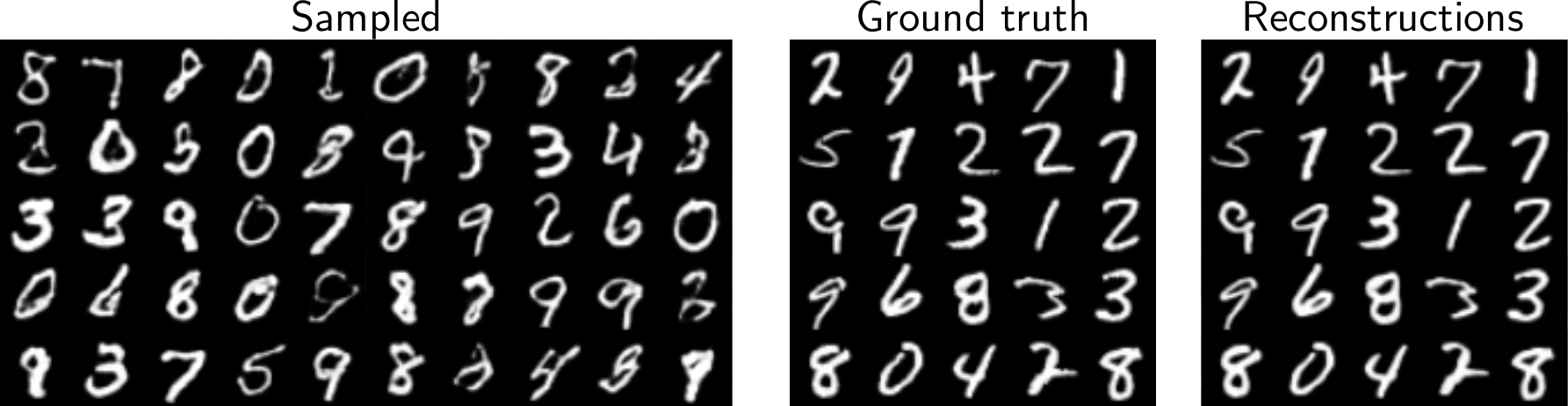}
    \caption{$\Relu$ $1\times1$ convolutions}
    \label{fig:mnist_relu_conv}
    \end{subfigure}
    \begin{subfigure}{0.9\textwidth}
    \includegraphics[width=\textwidth]{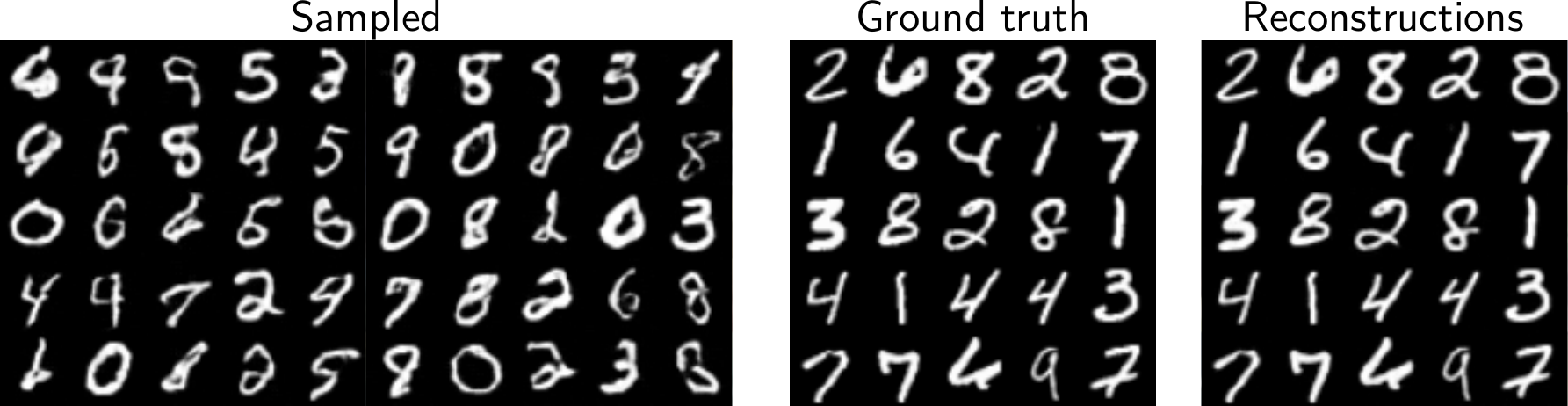}
    \caption{Linear $1\times1$ convolutions}
    \label{fig:mnist_linear_conv}
    \end{subfigure}
    \caption{{\inflow s} trained with (a) $\Relu$ and (b) linear $1\times1$ convolutions give similar sample quality.}
\end{figure*}

\begin{figure*}
    \centering
    \begin{subfigure}{0.90\textwidth}
    \includegraphics[width=\textwidth]{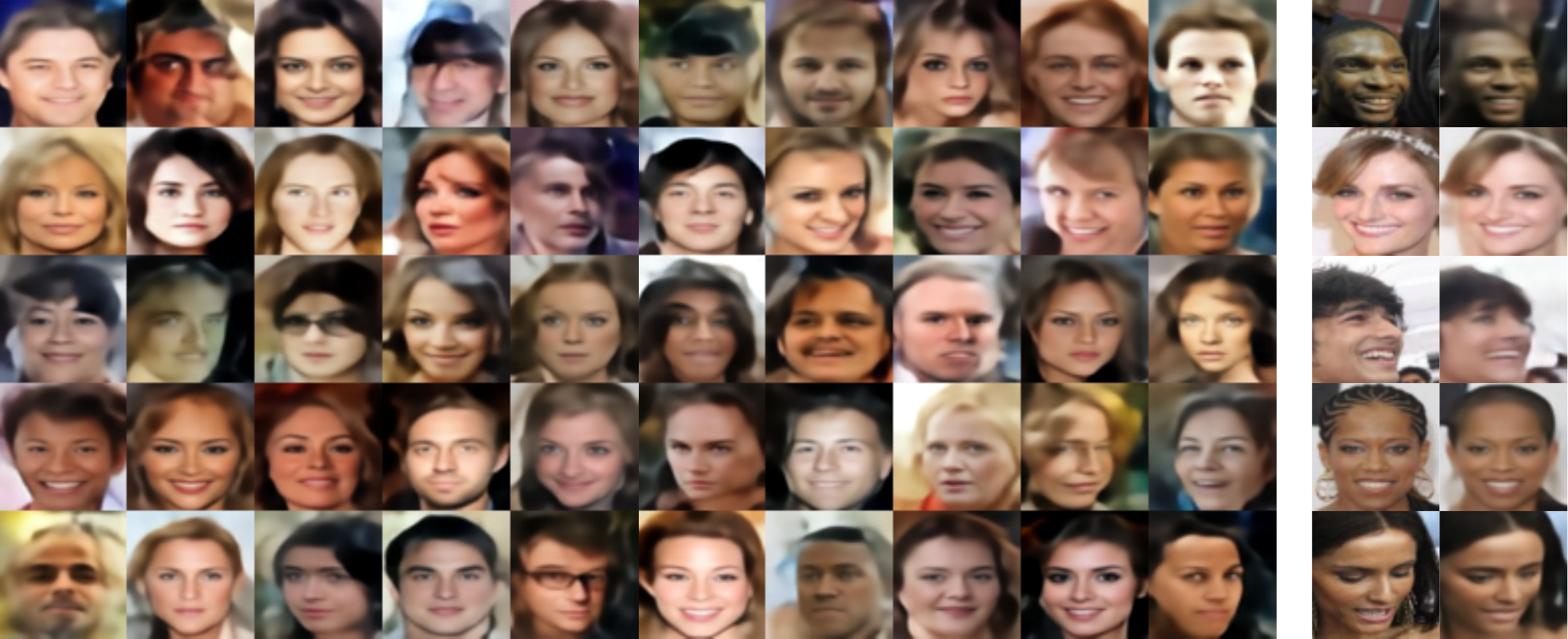}
    \caption{$\Relu$ $1\times1$ convolutions}
    \label{fig:celeba_relu_conv}
    \end{subfigure}
    \vfill
    \vfill
    \begin{subfigure}{0.90\textwidth}
    \includegraphics[width=\textwidth]{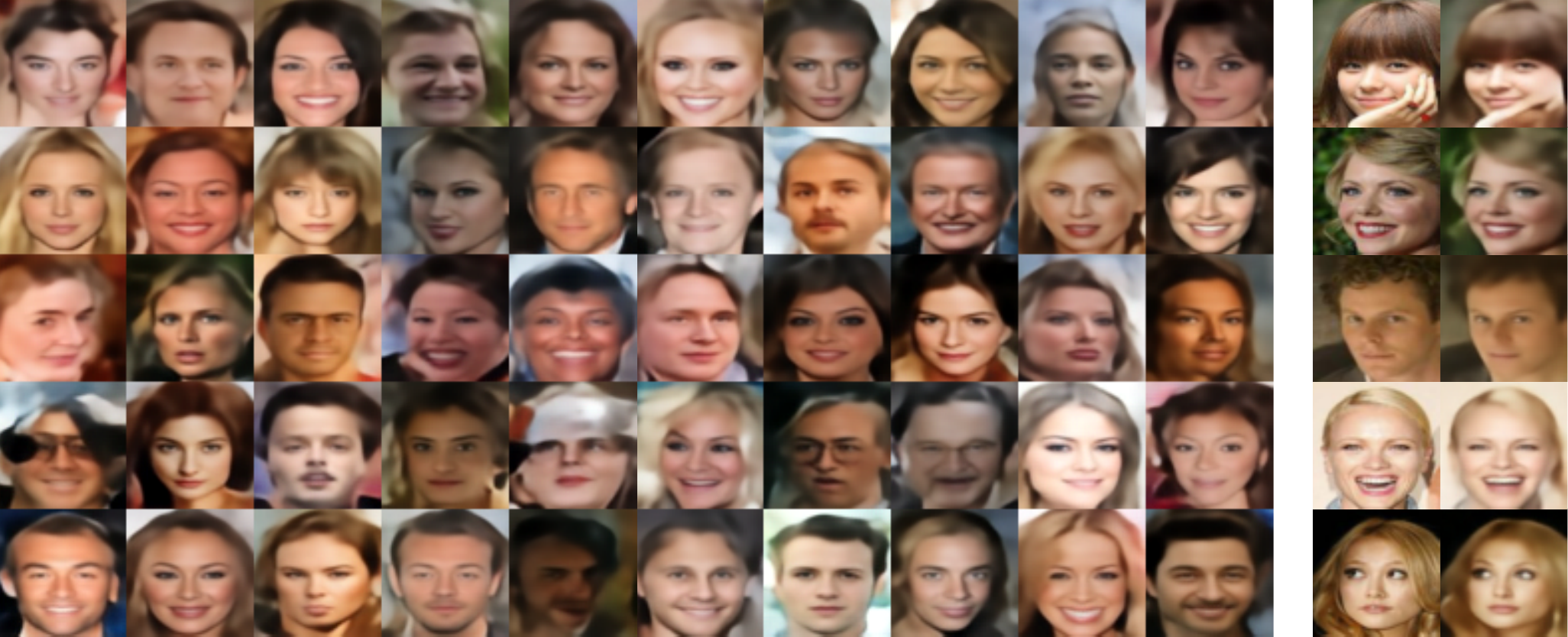}
    \caption{Linear $1\times1$ convolutions}
    \label{fig:celeba_linear_conv}
    \end{subfigure}
    \caption{{\inflow s} trained with (a) $\Relu$ and (b) linear $1\times1$ convolutions give similar sample quality. On the right, we showcase the reconstruction performance---the left column is ground truth and the right is our reconstruction (see Table \ref{tab:comp_resource} for quantitative assessment)}
\end{figure*}

\begin{figure*}
    \centering
    \includegraphics[width=0.9\textwidth]{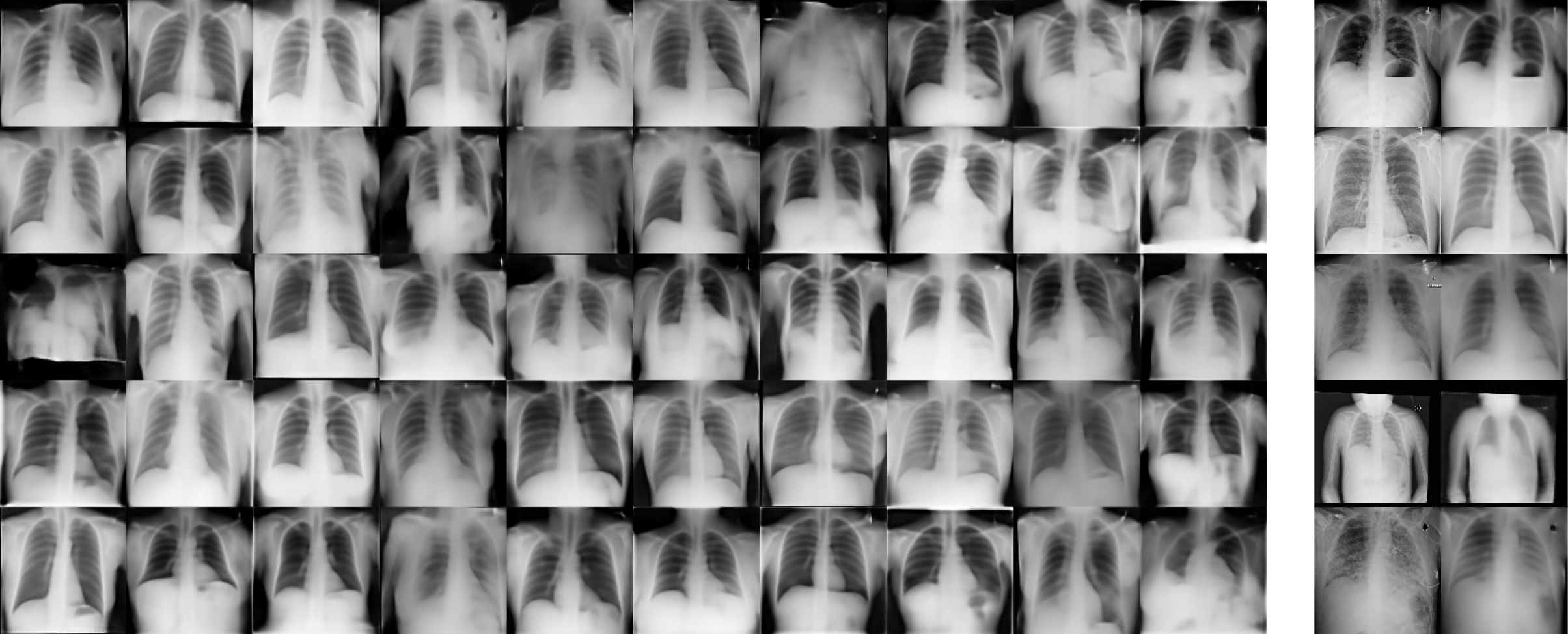}
    \caption{Generated samples on the Chest X-ray. On the right, we showcase the reconstruction performance---the left column is ground truth and the right is our reconstruction (see Table \ref{tab:comp_resource} for quantitative assessment) }
    \label{fig:chest_samples}
\end{figure*}

\begin{figure*}
    \centering
    \includegraphics[width=0.9\textwidth]{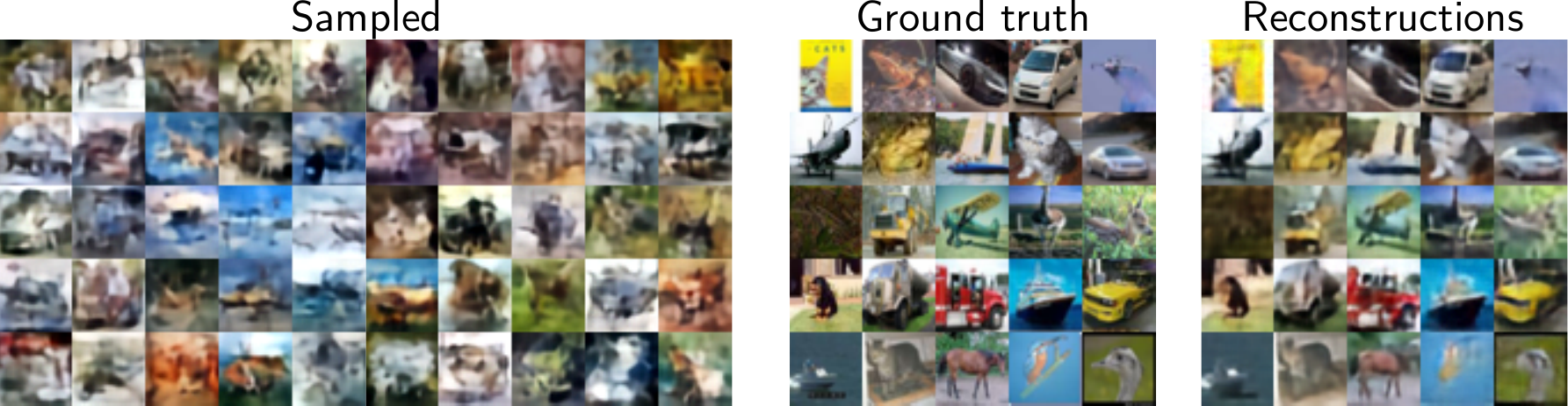}
    \caption{Generated samples and reconstructions of original data on the CIFAR-10 dataset.}
    \label{fig:cifar10_samples}
\end{figure*}

\begin{figure*}
    \centering
    \includegraphics[width= 0.9\textwidth]{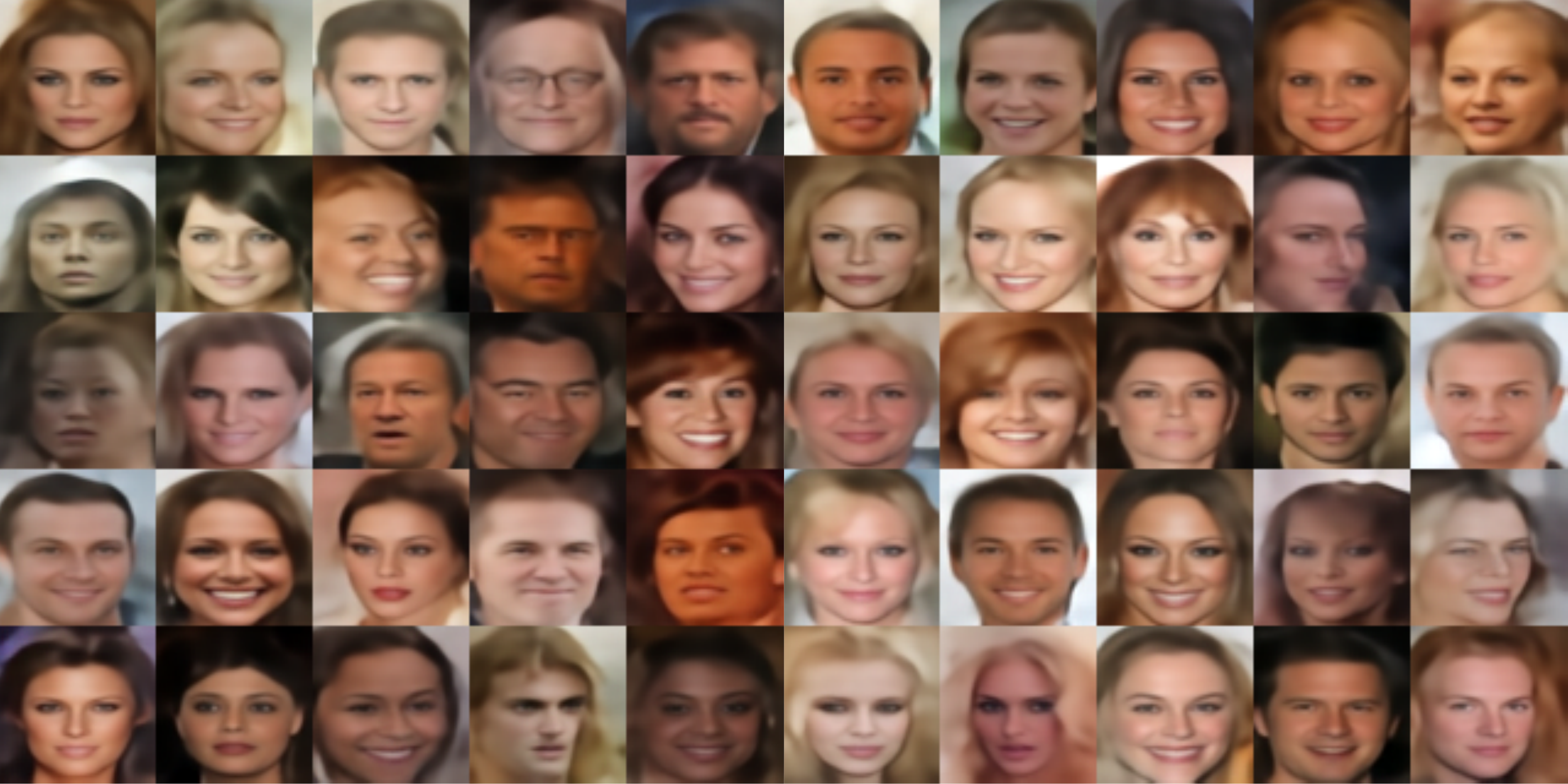}
    \caption{Generated samples on the celeba dataset with linear $1\times1$ convolution and $T = 0.85$.}
    \label{fig:celeba_linear_0.85}
\end{figure*}

% \begin{figure*}
%     \centering
%     \includegraphics[width=0.9\textwidth]{graphics/church_linaer_samples.pdf}
%     \caption{Generated samples and reconstructions of original data on the LSUN churches dataset.}
%     \label{fig:church_samples}
% \end{figure*}

%%%%%%%%%%%%%%%%%%%%%%%%%%%%%%%%%%%%%%%%%%%%%%%%%%%%%%%%%%%%%%%%%%%%%%%%%%%%%%%
%%%%%%%%%%%%%%%%%%%%%%%%%%%%%%%%%%%%%%%%%%%%%%%%%%%%%%%%%%%%%%%%%%%%%%%%%%%%%%%

\end{document}

% This document was modified from the file originally made available by
% Pat Langley and Andrea Danyluk for ICML-2K. This version was created
% by Iain Murray in 2018, and modified by Alexandre Bouchard in
% 2019 and 2021. Previous contributors include Dan Roy, Lise Getoor and Tobias
% Scheffer, which was slightly modified from the 2010 version by
% Thorsten Joachims & Johannes Fuernkranz, slightly modified from the
% 2009 version by Kiri Wagstaff and Sam Roweis's 2008 version, which is
% slightly modified from Prasad Tadepalli's 2007 version which is a
% lightly changed version of the previous year's version by Andrew
% Moore, which was in turn edited from those of Kristian Kersting and
% Codrina Lauth. Alex Smola contributed to the algorithmic style files.